%% file: main_decade.tex
\documentclass[10pt]{article} 
\usepackage{tmlr}

\input{math_commands.tex}

\usepackage{hyperref}
\usepackage{url}
\usepackage{titlesec}
\usepackage{algorithm}
\usepackage{algpseudocode}
\usepackage{wrapfig}

\usepackage{xcolor} 
\usepackage{tcolorbox} 

\usepackage{graphicx}
\usepackage{hyperref}

\usepackage{subcaption}
\usepackage{color}

\title{\centering{A Decade of Deep Learning:\\ A Survey on The Magnificent Seven}}

\author{
    \centering 
    Dilshod Azizov$^{1}$, Muhammad Arslan Manzoor$^{1}$, Velibor Bojković$^{1}$, \\
    Yingxu Wang$^{1}$, Zixiao Wang$^{1}$, Zangir Iklassov$^{1}$, Kailong Zhao$^{1}$, \\ Liang Li$^{1}$, Siwei Liu$^{1}$,
    Yu Zhong$^{2}$, Wei Liu$^{2}$ \& Shangsong Liang$^{2}$\thanks{Corresponding author.} \\
    $^{1}$\texttt{Mohamed bin Zayed University of Artificial Intelligence, UAE} \\
    $^{2}$\texttt{Sun Yat-sen University, China} \\
    \texttt{\{dilshod.azizov, muhammad.arslan\}@mbzuai.ac.ae}, \texttt{liangshangsong@gmail.com}
}

\begin{document}

\maketitle

\begin{abstract}
During the past decade\footnote{We consider the period from 2013 to 2024 due to the introduction and relevance of transformative architectures such as VAEs (2013), GANs (2014), ResNets (2015), GNNs (2016), Transformers (2017), Diffusion Models (2020), and CLIP (2021). Extending the timeframe to 2024 allows for a comprehensive and up-to-date analysis of these models, emphasizing their lasting impact and continued evolution in AI.}, deep learning algorithms have revolutionized the field of artificial intelligence (AI), leading to significant advancements in various domains. At the core of this transformation is the development of multi-layered neural network architectures that facilitate automatic feature extraction from raw data, significantly improving the efficiency on machine learning tasks. Given the rapid pace of these advancements,  an accessible manual is necessary to distill the key advances of the past decade. With this in mind, we introduce a study which highlights the evolution of deep learning, largely attributed to powerful algorithms. Among the multitude of breakthroughs, certain algorithms, including \emph{Residual Networks (ResNets),  Transformers, Generative Adversarial Networks (GANs), Variational Autoencoders (VAEs), Graph Neural Networks (GNNs), Contrastive Language–Image Pre-training (CLIP) and Diffusion models,} have emerged as the cornerstones and driving forces behind the discipline. We select these algorithms via a survey targeting a broad spectrum of academics and professionals with the aim of encapsulating the essence of the most influential algorithms over the past decade. In this work, we provide details on the selection methodology, exploring the mentioned architectures in a broader context of the history of deep learning. We present an overview of selected core architectures, their mathematical underpinnings, and the algorithmic procedures that define the subsequent extensions and variants of these models, their applications, and their challenges and potential future research directions. In addition, we explore the practical aspects related to these algorithms, such as training and optimization methods, normalization techniques, and rate scheduling strategies that are essential for their effective implementation. Therefore, our manuscript serves as a practical survey for understanding and applying these crucial algorithms and aims to provide a manual for experienced researchers transitioning into deep learning from other domains, as well as for beginners seeking to grasp the trending algorithms.

\end{abstract}

\section{Introduction}
\label{sec:introduction}

In the field of artificial intelligence (AI), deep learning (DL) algorithms have advanced significantly in the last decade, paving the way for advances in other areas~\cite{shrestha2019review} such as Computer Vision (CV), Natural Language Processing (NLP), Speech Recognition (SR), Robotics, etc. The core of deep learning, which is built on multi-layered neural network architectures, has caused a paradigm change by making it easier to automatically extract features and patterns from unprocessed data, thus improving the efficiency of machine learning tasks~\cite{najafabadi2015deep}. Further, the study by~\citet{lecun2015deep} shows the transformative potential of deep learning in various domains, from NLP to CV. Furthermore, advances in computational power and algorithmic efficiency allow deep learning neural networks to achieve remarkable accuracy, significantly outperforming traditional methods in many tasks, largely due to the development of strong algorithms, the growth of large datasets, and the availability of powerful computing resources~\cite{schmidhuber2015deep, sejnowski2018deep, wang2023enhancing}.

Out of all the algorithms, some (e.g., transformers)~\cite{vaswani2017attention, devlin2018bert, brown2020language} illustrated exceptional applicability and the advancement in the discipline. They have promoted better comprehension and handling of complicated high-dimensional data, particularly in the field of NLP, where the complexity of spoken language presented enormous obstacles~\cite{tsvetkov2016linguistic, radford2019language, lin2022survey}. Other examples of such impactful algorithms include ~\cite{nguyen2024image, carion2020end} (in the domain of CV), ~\cite{10.1145/3617833, ramachandran2017deep, kim2021vilt} (in the domain of multi-modal learning), and ~\cite{esteva2017dermatologist, rajpurkar2020chexnet} (in the domain of medical imaging).

In this study, we focus on the following seven algorithms,  aiming to distill the core ideas that have defined the most influential advancements in deep learning over the past decade. The algorithms we choose for our study are: Transformers~\cite{vaswani2017attention}, which revolutionized sequence-to-sequence tasks;  Residual Networks (ResNets)~\cite{he2016deep} which enabled deeper neural networks; Variational Autoencoders (VAEs)~\cite{kingma2013auto}, where probabilistic inference is a key component; Generative Adversarial Networks (GANs)~\cite{goodfellow2014generative}, which generate realistic data; Graph Neural Networks (GNNs)~\cite{kipf2016semi}, which extended neural networks to graph-structured data;  Contrastive Language–Image Pretraining (CLIP)~\cite{radford2019language}, which aligned vision and language representations; and Diffusion models~\cite{ho2020denoising}, which model complex distributions. The selection process for these algorithms was laborious and began with a survey sent to a large number of respondents (e.g., Bachelor, M.Sc. and Ph.D. students, research assistants, postdoctorals, and faculty members) to find out which DL algorithms they thought were most helpful and influential. Following a survey of experts, twelve algorithms were initially identified. We subsequently curated a list of seven algorithms that have emerged in the past ten years and remain highly relevant, while excluding five others that have become outdated (e.g. CNN, LSTM). Next, we rank each algorithm according to its impact on the research community as measured by the citation score of the subsequent core articles from \emph{Google Scholar}\footnote{Here is the ranking of algorithms based on the citation scores of their foundational papers (as of December 10, 2024): 1) Residual Networks (ResNets): ``Deep Residual Learning for Image Recognition'' by He et al. (2015) – 247,440 citations. 2) Transformers: ``Attention Is All You Need'' by Vaswani et al. (2017) – 144,501 citations. 3) Generative Adversarial Networks (GANs): ``Generative Adversarial Nets'' by Goodfellow et al. (2014) – 75,445 citations.
4) Variational Autoencoders (VAEs): ``Auto-Encoding Variational Bayes'' by Kingma and Welling (2013) – 40,549 citations.
5) Graph Neural Networks (GNNs): ``Semi-Supervised Classification with Graph Convolutional Networks'' by Kipf and Welling (2016) – 40,130 citations.
6) Contrastive Language-Image Pre-training (CLIP): ``Learning Transferable Visual Models From Natural Language Supervision'' by Radford et al. (2021) – 25,847 citations. 7) Diffusion Models: ``Denoising Diffusion Probabilistic Models'' by Ho et al. (2020) – 15,632 citations.}.

\begin{tcolorbox}[
    colback=cyan!5, 
    colframe=cyan!100, 
    title=\textbf{Why this Survey?}, 
    fonttitle=\bfseries, 
    sharp corners, 
    boxrule=1mm, 
    width=\textwidth 
    ]

To our knowledge, our manuscript is the first to survey the top seven deep learning algorithms of the last decade, providing a much-needed practical resource for understanding and applying these algorithms. By distilling the key concepts and advancements in this field, we aim to create a manual for experienced researchers transitioning into deep learning from other domains, as well as for beginners seeking to grasp the past decade trending algorithms - a crucial skillset in todays' technology-driven landscape.

\end{tcolorbox}

\textbf{Our contributions are as follows:}
\begin{enumerate}
    \item We discuss the evolution of deep learning and our methodology for selecting the top seven deep learning algorithms of the past decade.
    \item We offer an analysis of each algorithm, by covering core architecture, mathematical foundations, algorithmic procedure, training and optimization, extensions and variants, practical applications, challenges and future potential research directions, and summary.
\end{enumerate}

\begin{figure}[!t]
  \centering
  \includegraphics[width=0.85\linewidth]{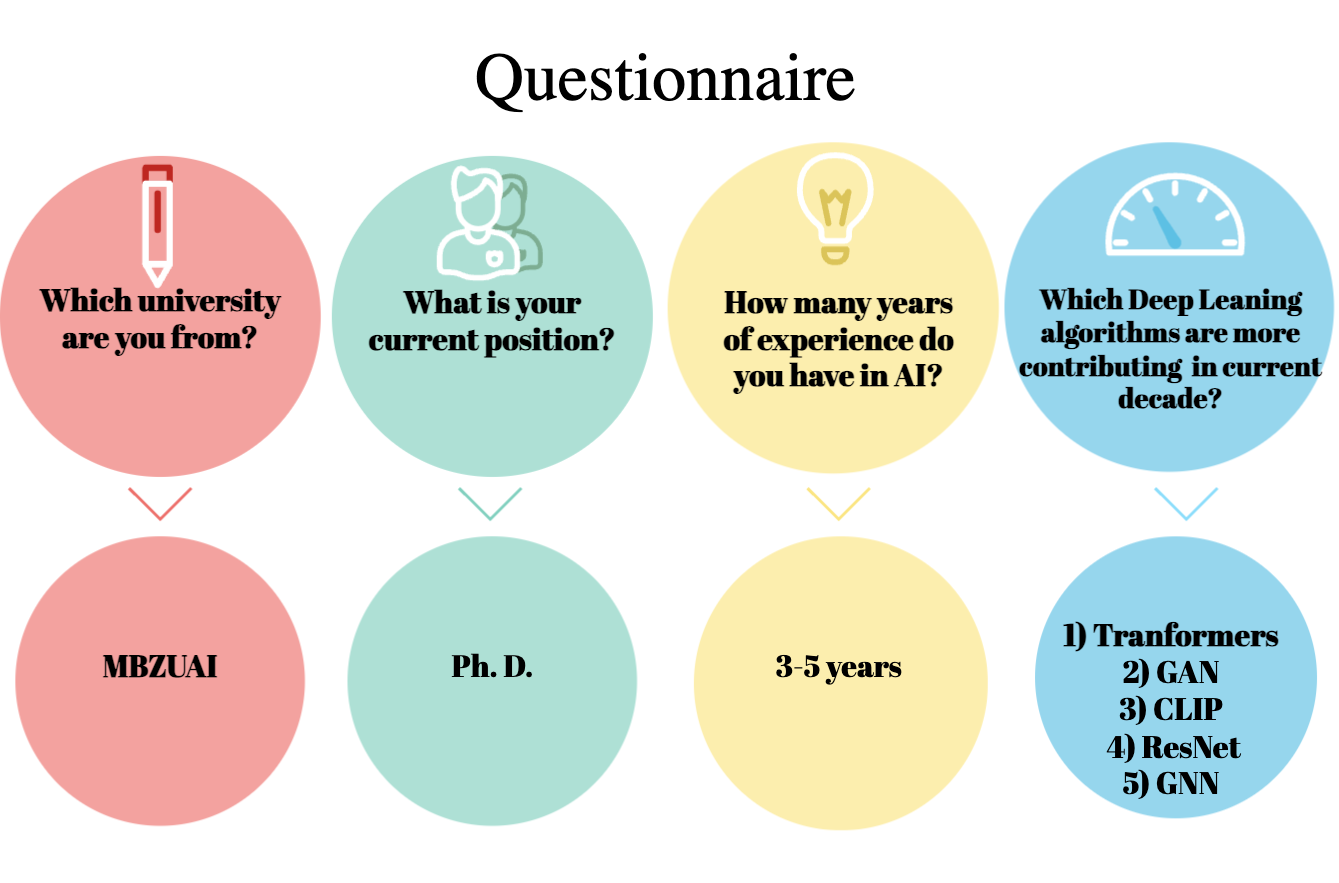}
  \caption{Examples of the questions provided to respondents, along with answers. The answers to the questions above have been randomly generated by the authors. For the question ``Which Deep Learning algorithm are more contributing in current decade?'' we suggested respondents to provide at least 5 algorithm examples.}
  \label{example}
\end{figure}

The contents of the paper are the following: In~\S~\ref{sec:evolution}, we analyze the chronological evolution of deep learning algorithms. In~\S~\ref{sec:methodology}, we explain our methodology for selecting the top seven algorithms in deep learning for the past decade, while in~\S~\ref{sec:algorithms}, we provide their detailed descriptions, both theoretical and practical. In~\S~\ref{sec:discussion}, we expand our discussion on basic building blocks, normalization techniques, optimization algorithms, and rate scheduling techniques, and we explore how these basic components are integral to the individual algorithms discussed in our study. Finally,~\S~\ref{sec:conclusion} summarizes our findings.

\section{Evolution of Deep Learning}
\label{sec:evolution}

The conceptual foundation for neural networks was laid in 1943 by~\citet{mcculloch1943logical} with the idea that neurons could be simulated with electrical circuits, proposing a model for artificial neurons. This early model demonstrated the potential to build computing systems that mimic biological processes. The subsequent development by~\citet{hebb1949first} introduced the concept of strengthening neural pathways through repeated use, laying the foundation for learning algorithms. In 1958~\citet{rosenblatt1958perceptron} introduced the perceptron, an early neural network model capable of simple pattern recognition. This model had two layers of processing units and demonstrated early success in tasks such as image and pattern recognition~\cite{rumelhart1986learning}. However, the initial excitement was dampened by~\citet{minsky1969introduction} who illustrated the perceptron inability to solve simple XOR functions, highlighting its limitations and ushering in a period of reduced interest and funding, known as the AI winter. A major breakthrough came with the introduction of the backpropagation algorithm in the 1970s~\cite{linnainmaa1970representation}. Its implications for deep neural network training were not fully appreciated and utilized until later~\cite{rumelhart1986learning, lecun1988theoretical}. This algorithm enabled the training of multi-layer networks and facilitated the learning of complex datasets by adjusting the weights in response to errors, revitalizing the interest in neural networks~\cite{hinton1986learning, schmidhuber2015deep}. The late 1980s and 1990s saw the development of various specialized neural network architectures. The neocognitron, introduced in 1988, was a hierarchical neural network that greatly improved visual pattern recognition~\cite{fukushima1988neocognitron}. The introduction of long-short-term memory (LSTM) networks and their subsequent applications underscored the ability of deep learning to handle not only single data points, but also entire data sequences~\cite{hochreiter1997long, gers1999learning}. Following this, Yann LeCun’s work in 1998 on convolutional neural networks (CNNs) with backpropagation marked significant progress in document analysis and image processing~\cite{lecun1998gradient}. Additional studies in this period contributed to the refinement and widespread application of these architectures~\cite{krizhevsky2012imagenet, simonyan2014very}.

The term ``Deep Learning'' was popularized in the 2000s, particularly with the development of Deep Belief Networks (DBNs)~\cite{hinton2006fast}. These networks used a layer-wise pre-training technique that significantly improved the efficiency of deep network training~\cite{hinton2006fast, bengio2007greedy}. This period also witnessed the increasing use of General-Purpose Graphics Processing Units (GPGPUs), which provided the computational power necessary to train complex models on large datasets~\cite{schmidhuber2015deep, raina2009large}. By 2012, deep learning began to dominate AI, driving progress in various fields, including SR, NLP, and CV. This era also marked the rise of large-scale data (big data), which further accelerated deep learning research and applications~\cite{kaisler2013big}.

One notable example of applications of these developments is Google AlphaGo, which demonstrated impressive deep learning capabilities. At the beginning of 2017~\cite{silver2017mastering}, the computer program, under the pseudonym ``Master,'' won three consecutive online games against professional human Go players, including a remarkable victory over Ke Jie. Using cutting-edge deep learning algorithms and extensive hardware resources, AlphaGo showcased its ability to defeat world-champion Go players~\cite{lee2018ai}.

\begin{figure}[!t]
    \centering
    \begin{subfigure}[b]{0.365\linewidth}
        \includegraphics[width=\linewidth]{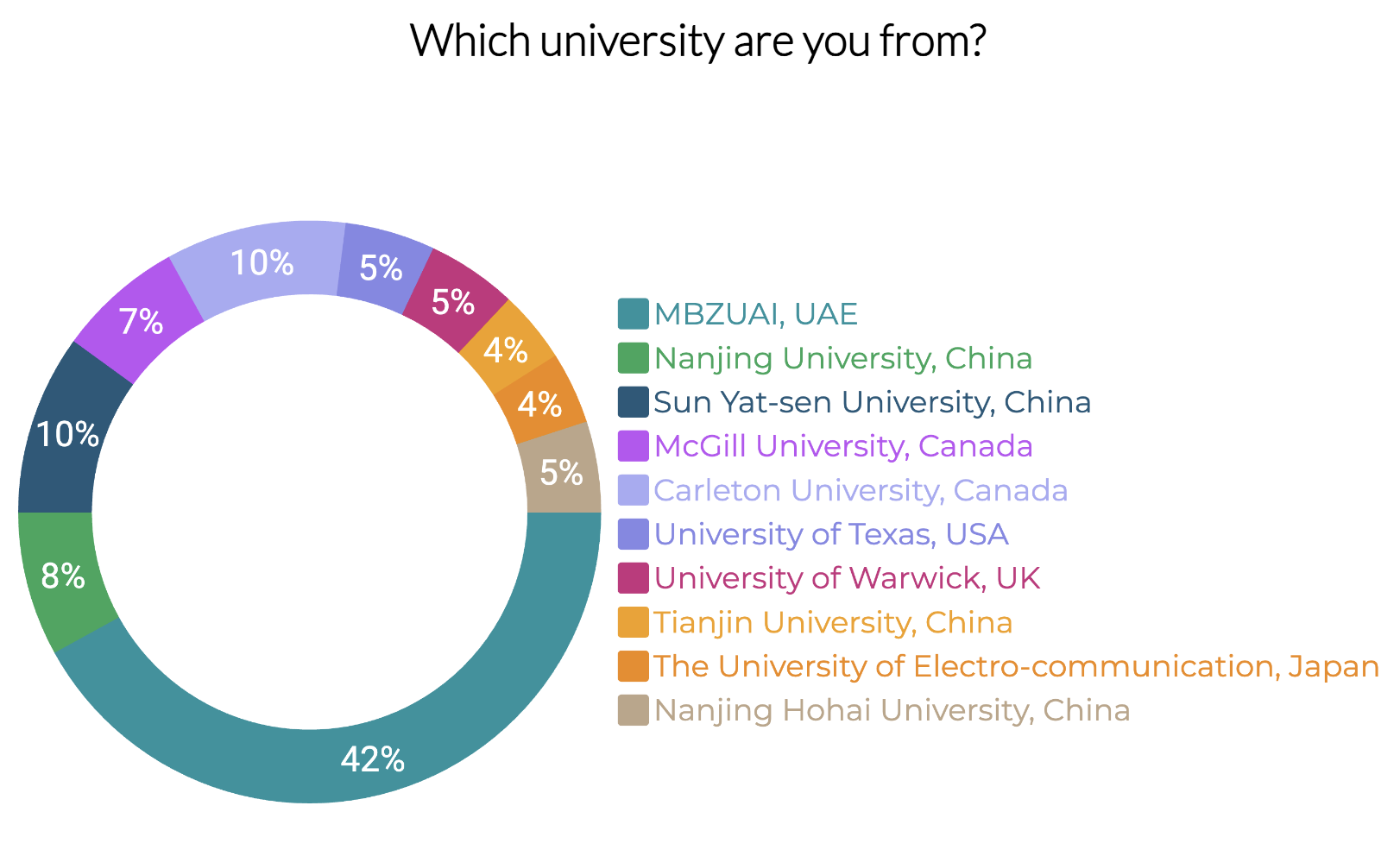}
        \caption{Distribution of the respondents across universities and countries.}
        \label{fig:university}
    \end{subfigure}
    \hfill
    \begin{subfigure}[b]{0.31\linewidth}
        \includegraphics[width=\linewidth]{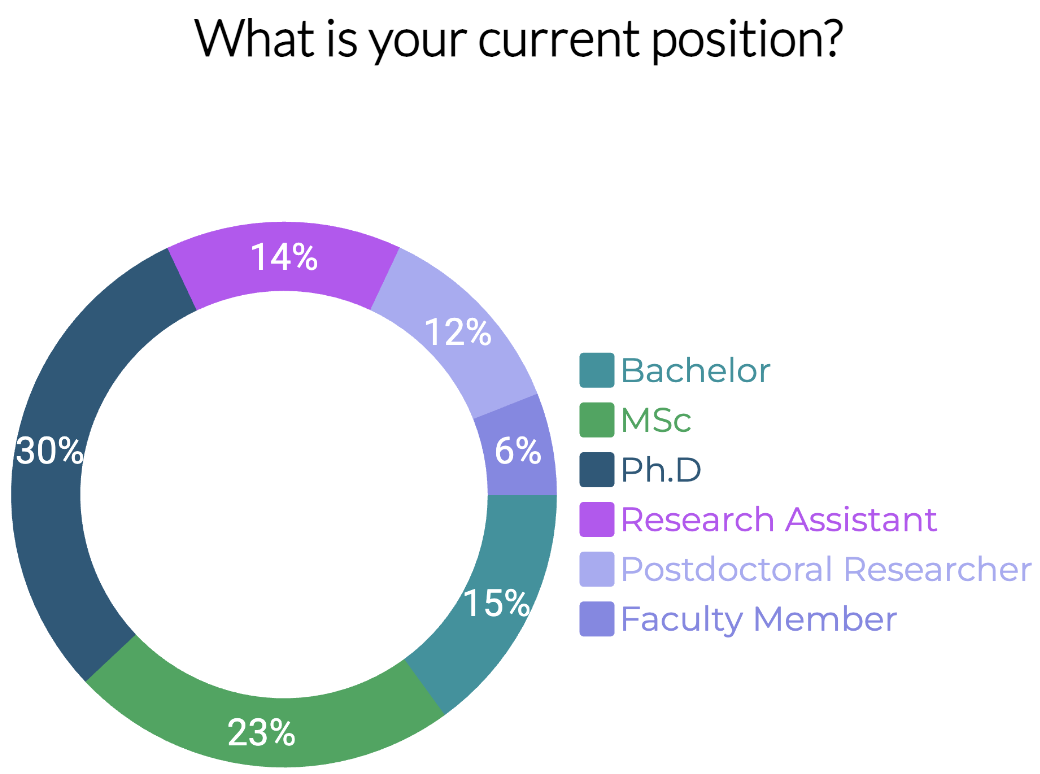}
        \caption{Distribution of the respondents based on positions.}
        \label{fig:position}
    \end{subfigure}
    \hfill
    \begin{subfigure}[b]{0.31\linewidth}
        \includegraphics[width=\linewidth]{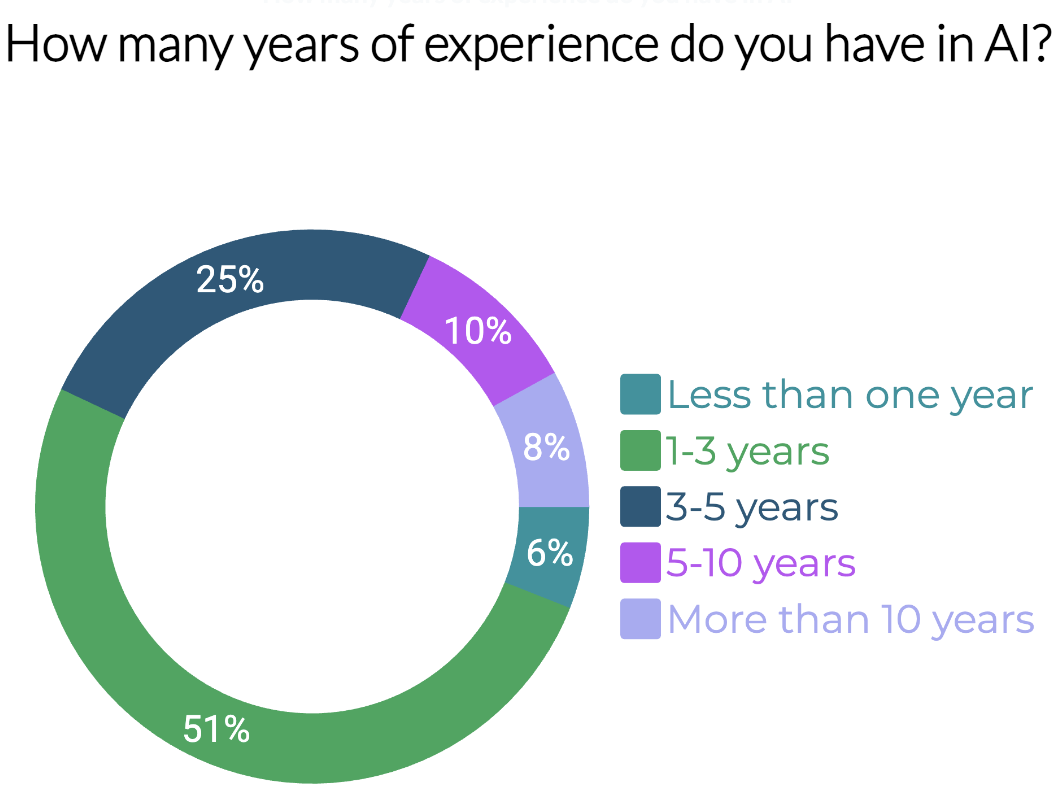}
        \caption{Distribution of the respondents based on experience.}
        \label{fig:experience}
    \end{subfigure}
    \caption{Distribution of respondents in our survey questionnaire.}
\end{figure}

Deep learning has revolutionized various fields, including NLP, CV, and beyond, through the introduction of innovative architectures such as ResNets, Transformer, GANs, VAEs, GNNs, CLIP, diffusion models. These advances have allowed the development of groundbreaking models such as BERT~\cite{devlin2018bert} and the GPT series~\cite{radford2018improving, brown2020language}, which significantly improved language understanding and generation capabilities. The emergence of efficient techniques, such as AutoML and neural architecture search, has further simplified model design and deployment processes. More recently, few-shot learning has gained prominence, with models like GPT-4o~\cite{achiam2023gpt} and Llama 3.1 (8B, 70B and 405B)~\cite{dubey2024llama} showcasing remarkable generalization abilities from limited examples. Additionally, there is an increasing focus on sustainable and efficient training methods to mitigate the environmental impact of large-scale model training~\cite{schwartz2020green}. The integration of deep learning into edge computing has also broadened its applications, enabling AI accessibility on a variety of devices~\cite{zhou2019edge}.

Deep learning has evolved from foundational models to sophisticated systems, driving transformative impacts across industries. This growth is fueled by global collaboration and ongoing innovations~\cite{smith2020quantum}. Ethical considerations, including data bias and privacy, remain critical and require responsible AI practices~\cite{doe2021ethical}.

\section{Methodology}
\label{sec:methodology}
The methodology used in this study is designed to ensure a robust and credible examination of the algorithms selected within the context. The process is divided into distinct phases, including data collection, analysis, and algorithm selection, each of which is delineated below in~\S~\ref{subsec:criteria}.
\subsection{Selection Criteria for the Seven Algorithms}
\label{subsec:criteria}

This study began by selecting the top algorithms that have been significantly influenced over the past decade through a survey designed to gather insights from a diverse group of experts in AI and computer science. The questionnaire included four questions (see Figure~\ref{example}). Figure~\ref{fig:university} displays the distribution of participants in universities and countries, with a majority of contributors from Mohamed bin Zayed University of Artificial Intelligence (MBZUAI) and Sun Yat-sen University. Figure~\ref{fig:position} illustrates that most of the respondents are Ph.D. and M.Sc. students. Figure~\ref{fig:experience} shows the levels of experience, with most having 1-3 years and 3-5 years of expertise. The total group of 100 respondent consisted of faculty, research assistants, postdoctoral researchers, and students at various educational stages, ensuring a breadth of perspectives by distributing the survey across multiple academic institutions.

Participants were asked to provide the five most significant algorithms in their view. Following the aggregation of responses, the top 12 algorithms were determined based on the collective vote of the participants. The next step was to carefully investigate the chronological beginnings of these algorithms in order to ascertain that they were created during the past decade. Several algorithms were eliminated as a result of this examination because the algorithms mentioned by the respondents were not introduced in our considered timeline (e.g., CNN and LSTM), leaving a final list of our seven algorithms -- \textbf{The Magnificent Seven}.

The second phase of the methodology went to an in-depth examination of the selected algorithms. The research impact of each algorithm was measured using the citation score of the seminal articles, obtained from a reputable academic database \emph{Google Scholar}. This measure served as a proxy for the algorithm's influence and acceptance within the community. Finally, we introduce each algorithm based on its citation score from \emph{Google Scholar} and introduce them accordingly in our study.

\section{Algorithms}
\label{sec:algorithms}
This section explores seven influential deep learning algorithms\footnote{Here are some source codes where we can find examples and implementations of the introduced standard algorithms, which we can easily load and use for the variety of tasks (by simply clicking on the source names provided after each algorithm in the brackets): Transformers \emph{\href{https://huggingface.co/docs/transformers/v4.17.0/en/index}{(Hugging Face Transformers}} and \emph{\href{https://github.com/tensorflow/models/tree/master/official/nlp}{TensorFlow Model Garden)}},  VAEs \emph{\href{https://github.com/keras-team/keras-io/blob/master/examples/generative/vae.py}{(Keras)}}, ResNets (\emph{\href{https://pytorch.org/hub/pytorch_vision_resnet/}{PyTorch}}), GANs \emph{\href{https://www.tensorflow.org/tutorials/generative/dcgan}{(TensorFlow}} and \emph{\href{https://pytorch.org/tutorials/beginner/dcgan_faces_tutorial.html}{PyTorch)}}, GNNs \emph{\href{https://github.com/pyg-team/pytorch_geometric}{(PyTorch Geometric)}}, CLIP \emph{\href{https://github.com/openai/CLIP}{(OpenAI}} and \emph{\href{https://huggingface.co/docs/transformers/en/model_doc/clip}{Hugging Face)}}, and Diffusion models \emph{\href{https://github.com/openai/guided-diffusion}{(OpenAI}} and \emph{\href{https://huggingface.co/docs/diffusers/en/tutorials/basic_training}{Hugging Face)}}.} - ResNets, Transformer, GANs, VAEs, GNNs, CLIP, Diffusion Models - that have reshaped the landscape of AI research and applications in NLP, CV, and generative modeling.

\subsection{ResNets}
\label{sec:resnet}

\textbf{\underline{Overview.}} Introduced by~\citet{he2016deep} Residual Networks (ResNets) represent a groundbreaking development in deep learning architectures. This architecture addresses the problem of training very deep neural networks by using skip connections or shortcuts to jump over some layers. Typical ResNets are composed of repeated blocks that have these residual connections, which allows training of networks that are substantially deeper than those used previously~\cite{he2016identity}. This design mitigates the problem of vanishing gradients by allowing the gradient to be directly backpropagated to earlier layers. The effectiveness of ResNets has been demonstrated in a broad range of applications, from image recognition~\cite{he2016deep} and object detection~\cite{ren2015faster} to semantic segmentation~\cite{chen2017rethinking} and medical image analysis~\cite{litjens2017survey}. ResNets have profoundly impacted the field, pushing the boundaries of what deep learning models can achieve. Recent developments have expanded the original ResNet by integrating new normalization techniques and activation functions to further improve performance and efficiency~\cite{zagoruyko2016wide, xie2017aggregated}. These advancements have made ResNets highly influential not only in academic research but also in practical deployments across various industries, including healthcare, automotive, and entertainment, reflecting their wide applicability and transformative potential in the field.

\begin{wrapfigure}[28]{r}{0.3\textwidth}
    \centering
    \includegraphics[width=\linewidth]{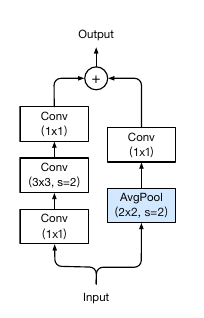}
    \caption{The ResNet architecture uses residual blocks with identity mappings, enabling deeper networks by addressing vanishing gradients and allowing efficient feature learning through shortcut connections.} 
    \label{ResNET}
\end{wrapfigure}

\textbf{\underline{Core Architecture.}} The ResNet architecture (see Figure~\ref{ResNET}) is designed to allow the training of extremely deep neural networks that can have hundreds or even thousands of layers efficiently. This is achieved through the use of residual blocks, which are the fundamental components of the ResNets architecture. The architecture can be described as follows:

\begin{itemize}
    \item \emph{Residual Blocks:} Each residual block in a ResNets contains two main paths. The first path is the weight layer path, typically consisting of two or three convolutional layers depending on the variant (e.g., ResNet-50, ResNet-101). These layers are followed by batch normalization layers and ReLU activation functions. The second path is the shortcut connection that skips these layers.
    
    \item \emph{Shortcut Connections:} The key innovation in ResNets is the shortcut connection that skips one or more layers. Shortcut connections simply perform identity mapping, and their outputs are added to the outputs of the stacked layers. This design addresses the vanishing gradient problem by allowing gradients to flow directly through the network during the backward pass.

\end{itemize}

This architectural innovation enables ResNets to achieve higher performance and faster training with deeper networks, leading to better performance in a variety of tasks such as image classification, object detection, etc.

\textbf{\underline{Mathematical Foundations.}} ResNets introduce a fundamental shift in how layer inputs and outputs are handled in deep neural networks. 

\emph{Residual Learning Function.} The key innovation in ResNets is the residual learning function, which is designed to make it easier to optimize deeper networks. The function is mathematically formalized through the following equation:

\begin{equation}
    \mathbf{y} = \mathcal{F}(\mathbf{x}, \{W_i\}) + \mathbf{x}\,.
\end{equation}

\noindent Here, \( \mathbf{x} \) and \( \mathbf{y} \) are the input and output of the layers considered. The function \( \mathcal{F}(\mathbf{x}, \{W_i\}) \) represents the residual mapping to be learned. For each layer, instead of trying to learn an underlying mapping directly, ResNets learns the difference \( \mathcal{F} \) between the input and output, which is theoretically easier to optimize, especially in very deep networks.

\emph{Identity Shortcut Connection.} The addition of the shortcut connection \( \mathbf{x} \) to the output of the residual function \( \mathcal{F} \) helps to address the problem of vanishing gradients by allowing gradients to flow directly through the identity function:

\begin{equation}
    \frac{\partial \mathcal{L}}{\partial \mathbf{x}} = \frac{\partial \mathcal{L}}{\partial \mathbf{y}} \cdot (1 + \frac{\partial \mathcal{F}}{\partial \mathbf{x}})\,.
\end{equation}

\noindent This ensures that the gradient does not vanish quickly as it is propagated through many layers, making it possible to train networks with much greater depth.

\emph{Layer Normalization.} Each residual block in ResNets typically includes a batch normalization step after each convolutional layer. This normalization helps in stabilizing the learning process and reduces the sensitivity of the network to different initialization schemes.

\emph{Loss Functions and Variants.}
Standard ResNets typically use the cross-entropy loss function for classification tasks, but modifications and variants have been introduced, such as ResNets variants include pre-activation ResNets, modifying block operations for better training dynamics~\cite{he2016identity}, and ResNeXt, using grouped convolutions to boost capacity and efficiency without added complexity~\cite{xie2017aggregated} to address specific challenges.

\textbf{\underline{Algorithmic Procedure.}} The training of ResNets involves a systematic process that leverages the unique architectural features of residual blocks to facilitate the training of very deep networks. The following outlines the step-by-step procedure for training ResNets:

\begin{enumerate}
    \item \textbf{Input Processing:} Each input image is initially processed by a convolutional layer, which is typically followed by batch normalization and a ReLU activation function to prepare the data for subsequent layers.

    \item \textbf{Residual Block Processing:}
 
    \item \emph{Convolutional Layers:} Each residual block contains several convolutional layers. These layers apply a series of filters to the input data and are each followed by batch normalization and ReLU activation.
        
    \item \emph{Shortcut Connection:} Parallel to the convolutional layers, there is a shortcut connection that carries the input directly to the end of the residual block. This helps mitigate the vanishing gradient problem by allowing gradients to flow through the network without significant attenuation.
        
    \item \emph{Element-wise Addition:} The output of the last convolutional layer in the block and the shortcut connection are added element-wise. This combination is then passed through another ReLU activation function.

    \item \textbf{Repeat Block Processing:} The processed output of one residual block is then fed into the next. This sequence is repeated across all residual blocks in the network. Depending on the version of ResNets (e.g., ResNet-34, ResNet-50, ResNet-101), the depth and number of blocks will vary.

    \item \textbf{Final Layers:} After all residual blocks have processed the input data, the output is passed through a global average pooling layer, followed by a fully connected layer that acts as the classifier.
   \end{enumerate}

\textbf{\underline{Training and Optimization.}} Training ResNets effectively involves optimizing several key aspects to ensure model stability and high performance. Key to ResNets' training is the use of batch normalization, which normalizes the input layer by adjusting and scaling activations. This helps mitigate the internal covariate shift problem, leading to much faster convergence and stabilizing the training process across very deep networks~\cite{ioffe2015batch}. Other optimizations include:

\begin{itemize}
    \item \textbf{Residual Learning:} The design of shortcut connections greatly reduces the risk of vanishing gradients, facilitating the training of networks that are significantly deeper than previous architectures.
    \item \textbf{Learning Rate Scheduling:} Gradual adjustment of the learning rate, such as using a step decay, where the learning rate is reduced by a factor every few epochs, helps achieve a lower loss and better accuracy~\cite{he2016deep}.
    \item \textbf{Weight Initialization:} Careful initialization of weights, often using methods such as He initialization, which is tailored for layers followed by ReLU activations, supports the training of deep models by preventing the problem of exploding or diminishing gradient magnitudes~\cite{he2015delving}.
\end{itemize}

\textbf{\underline{Extensions and Variants.}}
Since its inception, various extensions and variants of ResNets have been developed to address specific needs and improve the efficiency of the architecture. These include ResNeXt, which incorporates grouped convolutions to increase the cardinality of the network without additional computational cost~\cite{xie2017aggregated}; Wide ResNets (WRN), which modifies the network by decreasing the depth and increasing the width, showing better performance and ease of training~\cite{zagoruyko2016wide}; Pre-activation ResNets, which improves the training dynamics and allows deeper models without degradation by modifying the original architecture to have batch normalization and ReLU activation before convolutions in residual blocks~\cite{he2016identity}; Bottleneck Design, used in deeper ResNets models, which reduces dimensionality and then restores it, reducing computational burden~\cite{he2016deep}; Efficient ResNets (EResNet), which achieves improved performance with reduced computational costs~\citet{xiong2024eresnet}.

\textbf{\underline{Applications.}} ResNets has been fundamental in the advancement of deep learning applications in various domains. The following are some of the prominent applications: Image Classification, where they have set new benchmarks and significantly reduced error rates compared to previous models \citep{he2016deep}; Object Detection, where they serve as the backbone for architectures like Faster R-CNN \citep{ren2015faster}; Semantic Segmentation, where they capture spatial hierarchies~\citep{chen2017rethinking}; Video Analysis, where they handle temporal data and capture dynamic patterns over time \citep{wang2016temporal}; Medical Imaging, where they analyze complex medical images to detect and diagnose diseases early and accurately \citep{shen2017deep}; and Feature Extraction, where they extract powerful features from images that can be used in various CV tasks beyond classification \citep{he2016deep}.

\textbf{\underline{Challenges \& Future Directions.}}
 \emph{Training and deployment} of ResNets face several key challenges that present opportunities for future research and development. Overfitting remains a persistent issue that can be addressed through advanced data augmentation techniques~\cite{shorten2019survey} and regularization methods like dropout and L2 regularization~\cite{srivastava2014dropout}. Future work could explore more robust regularization techniques and adaptive data augmentation strategies to further enhance model generalization.

The \emph{degradation problem,} caused by the decrease in performance in deeper networks has been mitigated with pre-activation in residual blocks~\cite{he2016identity} and bottleneck designs~\cite{he2016deep}. However, future studies could extend these solutions by employing Neural Architecture Search (NAS) to design task-specific, optimized ResNets.

 \emph{Scalability} remains challenging due to high computational and memory demands. Current efforts to optimize ResNets on hardware through mixed-precision training and distributed frameworks~\cite{micikevicius2017mixed} provide a foundation for future exploration of pruning, quantization, and knowledge distillation techniques. These approaches could make ResNets more efficient and adaptable to resource-constrained environments.

Furthermore, \emph{interdisciplinary applications} highlight the untapped potential of ResNets in fields such as genomics for trait prediction and climate modeling for improved forecasting~\cite{xu2020deep, rolnick2019tackling}. Finally, integrating ResNets with other deep learning paradigms, such as transformers or GANs, presents an exciting direction for creating hybrid architectures that take advantage of the strengths of multiple models.

\textbf{\underline{Summary.}} ResNets represent a transformative development in deep learning, fundamentally altering the approach to training very deep neural networks. Introduced by \citet{he2016deep}, ResNets effectively address the vanishing gradient problem through innovative skip connections, enabling the construction of networks that are both deeper and more performant than those previously possible \cite{he2016identity}. These networks have significantly advanced the field, setting new benchmarks in tasks such as image classification, object detection, and beyond, demonstrating robustness across a variety of applications \cite{ren2015faster}. Their architectural efficiency has not only improved the accuracy of the model, but also expanded the potential applications, impacting sectors from automotive to healthcare \cite{litjens2017survey}. However, challenges such as model scalability and overfitting underscore the need for ongoing research to refine and enhance ResNets architectures. Efforts continue to optimize their structure and expand their capabilities, ensuring that ResNets remain at the forefront of AI technology.

\subsection{Transformers}

\textbf{\underline{Overview.}} The introduction of the Transformer architecture by~\citet{vaswani2017attention} marked a significant evolution in the field of machine learning, particularly in the processing of sequential data. This architecture effectively addresses the limitations of previous sequence processing models, such as recurrent neural networks (RNNs)~\cite{rumelhart1986learning} and long-short-term memory networks (LSTMs)~\cite{hochreiter1997long}, by eliminating the need for sequential data processing. This innovation has significantly improved performance across a broad spectrum of applications, including NLP, CV \cite{nguyen2024image, carion2020end, wang2021pyramid, liu2021swin, arnab2021vivit, manzoor2020lexical, vayani2024all} and SR \cite{dong2018speech, karita2019comparative, li2019neural, tian2019self, miao2020transformer}, fundamentally changing our approach to model architecture in various domains.

\begin{wrapfigure}[25]{r}{0.3\textwidth}
  \centering
  \includegraphics[width=\linewidth]{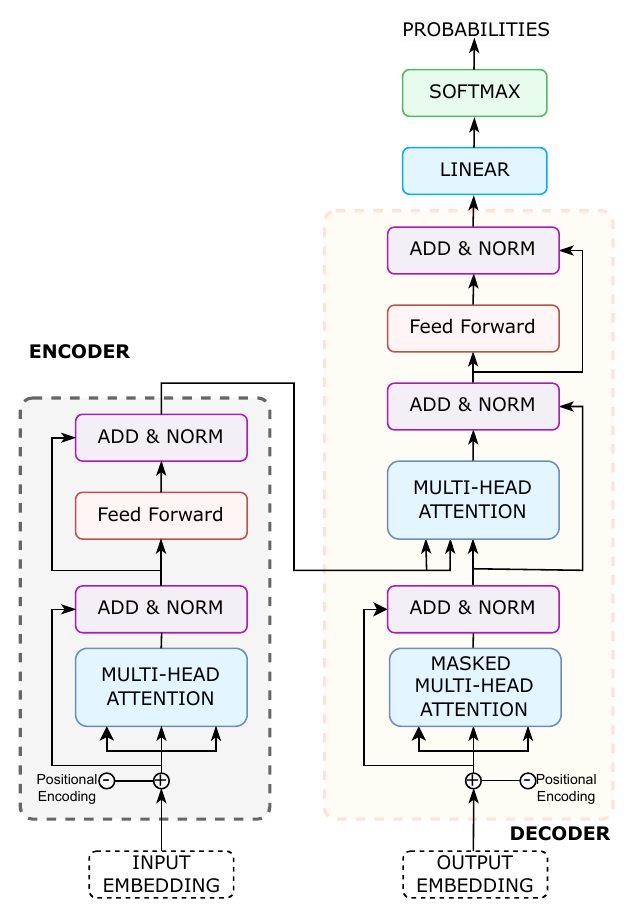}
  \caption{Illustration of the transformer architecture with encoder and decoder: the encoder processes input sequences, while the decoder generates outputs using self-attention, cross-attention, and feedforward layers.}
  \label{transformer}
\end{wrapfigure}

\textbf{\underline{Core Architecture.}} The transformer model eschews traditional recurrent layers and instead relies entirely on an attention mechanism to draw global dependencies between input and output. The architecture of a transformer is divided into two main parts as shown in Figure~\ref{transformer}: \emph{encoder} and \emph{decoder.}

The \emph{encoder} maps an input sequence of symbol representations \((x_1, ..., x_n)\) to a sequence of continuous representations \textbf{z}. Each layer of the encoder consists of two sub-layers: the first is a multi-head self-attention mechanism, and the second is a simple, position-wise fully connected feed-forward network. \citet{vaswani2017attention} introduce residual connections around each of the two sub-layers, followed by layer normalization.

The \emph{decoder} is also made up of a series of identical layers. In addition to the two layers in each encoder layer, each decoder layer has three sub-layers: a self-attention mechanism, an encoder-decoder attention mechanism, and a feed-forward neural network (FFNN). The encoder-decoder attention mechanism performs multi-head attention over the encoder's output. Similarly to the encoder, residual connections are added around each sub-layer, followed by layer normalization.

\textbf{\underline{Mathematical Foundations.}} The self-attention mechanism at the heart of the transformer uses scaled dot-product attention. The attention function can be described mathematically as mapping a query and a set of key-value pairs to an output, where the query, keys, values, and output are all vectors. The output is computed as a weighted sum of the values, where the weight assigned to each value is computed by a compatibility function of the query with the corresponding key. Specifically, attention weights are computed using the following equation:

\begin{equation}
\text{Attention}(Q, K, V) = \text{softmax}\left(\frac{QK^T}{\sqrt{d_k}}\right)V \,,
\end{equation}

\noindent where $Q$, $K$, and $V$ are the queries, keys, and value matrices, respectively, and $d_k$ is the dimension of the keys.

The transformer model, with its self-attention mechanism, excels in processing sequences with long-range dependencies. This architecture has improved language models and inspired adaptations in image recognition and speech processing.

\textbf{\underline{Algorithmic Procedure}}
The transformer model operates via a distinct sequence of steps within its \emph{encoder-decoder} architecture:
\begin{enumerate}
    \item Input Processing: Each input token is converted into a vector through embedding layers. Positional encodings are added to these embeddings to incorporate information about the position of each token within the sequence.

    \item {Encoder Layer Processing:}
    \begin{enumerate}
        \item \emph{Multi-Head Self-Attention:} The encoder uses self-attention mechanisms to process the input. This involves multiple attention heads computing attention scores simultaneously, allowing the model to capture different types of relationships between words in the input sequence.
        \item \emph{Add \& Norm:} The outputs from the self-attention layer is added to the original embeddings (residual connection) and normalized.
        \item \emph{Feed-Forward Networks:} The normalized output is then passed through a point-wise feed-forward network, which is applied to each position separately and identically.
        \item \emph{Add \& Norm:}  Another residual connection followed by layer normalization is performed after the feed-forward network.
    \end{enumerate}

    \item {Decoder Layer Processing:} Similar to the encoder, but with an additional step:
    \begin{enumerate}
        \item \emph{Masked Multi-Head Attention:} To prevent positions from attending subsequent positions, masked self-attention is used in the decoder.
        \item \emph{Encoder-Decoder Attention:} Attention mechanisms are used that focus on the output of the encoder stack, helping the decoder to focus on the appropriate parts of the input sequence.
        \item \emph{Feed-Forward Networks and Normalization:} As in the encoder, feed-forward and normalization steps are performed.
    \end{enumerate}
    
    \item Output Generation: The decoder output is transformed into predicted output tokens, typically through a linear transformation followed by a softmax layer to predict the probability of each token in the vocabulary.
\end{enumerate}

\textbf{\underline{Training and Optimization.}} Training the transformer model involves using the standard mini-batch gradient descent method combined with the Adam optimizer \cite{kingma2014adam}. The learning rate scheduling, crucial for stabilizing the training early on, is particularly specified as follows:

\begin{equation}
\text{lr} = H^{-0.5} \cdot \min(S^{-0.5}, S \cdot W^{-1.5}),
\end{equation}

\noindent
where: \( \text{lr} \): learning rate, \( H = d_{\text{model}} \): dimensionality of the model hidden layers, \( S = \text{step\_num} \): current training step number, \( W = \text{warmup\_steps} \): number of steps during the warm-up phase. This learning rate schedule combines the effects of the: inverse square root decay, controlled by \( S^{-0.5} \) and linear scaling during the warm-up phase, given by \( S \cdot W^{-1.5} \).

\textbf{\underline{Extensions and Variants.}} Since its inception, the transformer architecture has inspired a myriad of variants aimed at enhancing efficiency, scalability, and applicability across different modalities. The most notable among these are BERT (Bidirectional Encoder Representations from transformers) for natural language understanding tasks~\cite{devlin2018bert}, and GPT (Generative Pre-trained Transformer) for generative tasks~\cite{radford2018improving}. The adaptability of transformers has been extended to computer vision with the introduction of vision transformers (ViT)~\cite{nguyen2024image} and to video processing with Video Vision transformers (ViViT)~\cite{arnab2021vivit}. Further advancements include efficient transformers such as Linformer, which reduces the computational complexity significantly~\cite{wang2020linformer}, and the Performer, which provides an efficient approximation of self-attention for very long sequences~\cite{choromanski2020rethinking}. The emergence of cross-modal transformers, such as Perceiver, highlights the versatility of the transformer architecture, enabling it to handle various data types, including images, audio, and text~\cite{jaegle2021perceiver}. Similarly, emerging large language models (LLMs), such as GPT-4o1\footnote{Although the name GPT4o1 suggests the presence of transformer blocks in its architecture, the company behind this model, OpenAI, has not publicly disclosed detailed model architectures since the release of GPT-3.5.} and Llama3.1 (405B)~\cite{achiam2023gpt, dubey2024llama}, also demonstrate these capabilities. Furthermore, the application of transformers in the field of reinforcement learning exemplifies their utility in decision-making processes~\cite{chen2021decision, sun2024llm, deng2024novice}.

\textbf{\underline{Applications.}}
Transformers have become instrumental in various domains since their introduction, demonstrating their versatility and capability in numerous applications. These include natural language processing, where they have set new performance benchmarks in various NLP challenges (e.g., fake news detection)~\cite{azizov2024safari}, including language models such as GPT-4o1 and Llama 3.1 (8B, 70B and 405B). Additionally, transformers have improved the accuracy of speech recognition, exemplified by models such as the Conformer~\cite{gulati2020conformer}, HuBERT~\cite{hsu2021hubert} and Whisper~\cite{radford2022whisper, radford2023robust}. Furthermore, vision transformers (ViTs)~\cite{nguyen2024image} and applications in object detection with DETR~\cite{carion2020end} demonstrate that transformers can effectively process and understand visual data, rivaling and sometimes surpassing advanced CNN architectures.

\textbf{\underline{Challenges \& Future Directions.}}
Despite their widespread adoption, transformers face several challenges that provide key areas for future research and innovation. \emph{computational demand} is a major limitation due to the quadratic scaling of self-attention with sequence length~\cite{kitaev2020reformer}. Addressing this, future work is focusing on techniques like sparse attention mechanisms and advancements in specialized AI hardware to reduce computational and memory requirements.

\emph{Data efficiency} remains a hurdle, as transformers require vast amounts of data to train effectively, limiting their applicability in data-scarce scenarios~\cite{brown2020language, patil2024review}. Future directions include developing more data-efficient training strategies, such as transfer learning and semi-supervised learning, to enhance performance in low-data environments.

\emph{Interpretability} is another significant challenge, as transformers’ complex architectures make decision-making processes opaque~\cite{voita2019analyzing, singh2024rethinking}. Research into explainability frameworks aims to address this by making model decisions more transparent and justifiable, particularly in critical applications like healthcare and autonomous systems.

Finally, \emph{generalization} across tasks remains limited without extensive task-specific fine-tuning~\cite{jiang2020convbert, liu2024evaluating, zhang2024can}. Future work is exploring universal pretraining methods and modular architectures to enhance adaptability across diverse tasks and domains.

Emerging directions also include integrating transformers into \emph{multi-modal models} to unify text, image, and other data types, unlocking new applications in fields such as genetics and video processing. Furthermore, a focus on sustainable AI aims to reduce the environmental impact of training large-scale transformers through energy-efficient algorithms and carbon-conscious training practices.

\textbf{\underline{Summary.}}
Transformers have revolutionized machine learning with their powerful capabilities to handle sequence data in various applications. Introduced by \citet{vaswani2017attention}, this architecture has replaced recurrent layers with a more efficient self-attention mechanism, facilitating faster training and better handling of long-range dependencies. The transformers have not only established new standards in NLP, but have also expanded their influence to fields such as CV \cite{nguyen2024image} and SR \cite{gulati2020conformer}. 


\subsection{GANs}

\textbf{\underline{Overview.}} Generative Adversarial Networks (GANs), introduced by~\cite{goodfellow2014generative}, represent a transformative approach in machine learning, allowing the synthesis of highly realistic data in diverse domains. These networks comprise two competing models: a generator which attempts to generate data indistinguishable from genuine data, and a discriminator which aims to distinguish between real and generated data. This adversarial process improves through iterative training, where both models enhance their capabilities in a dynamic equilibrium~\cite{karras2019style, brock2018large}. Recent advances have expanded GAN applications from simple image generation to complex tasks such as super-resolution, style transfer, and synthetic data augmentation for training other machine learning models~\cite{wang2018high, choi2020stargan}. Furthermore, GANs have been pivotal in unsupervised learning, providing a powerful tool to discover intricate data distributions without labeled datasets~\cite{chen2016infogan}. Current research continues to address the challenges of training stability and mode collapse, with novel architectures such as BigGAN and StyleGAN offering substantial improvements in image quality and training methodology~\cite{brock2018large, karras2019style}.

\begin{figure}[!t]
    \centering
    \includegraphics[width=11cm,height=4.5cm]{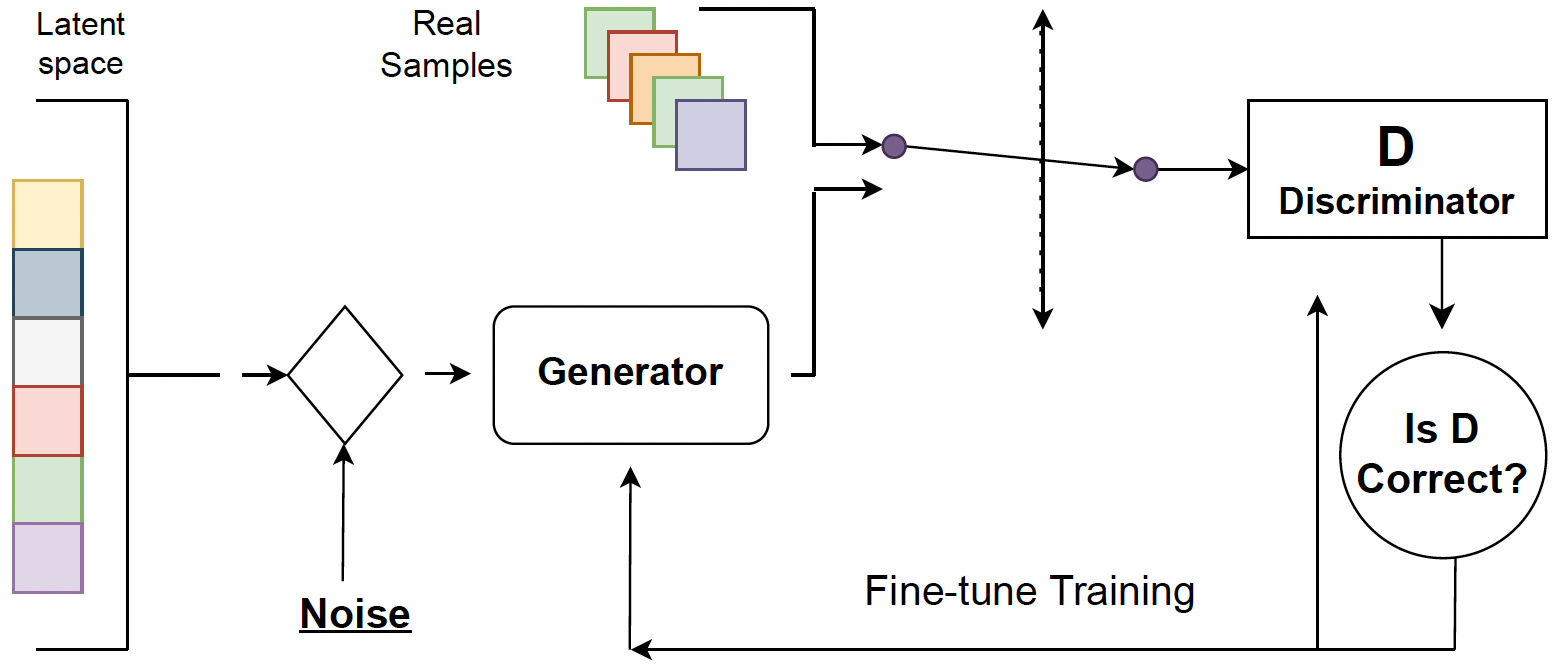}
    \caption{The GANs core architecture involves two neural networks: a generator creating fake data, and a discriminator distinguishing real from fake, trained adversarially.} 
    \label{GANs}
\end{figure}

\textbf{\underline{Core Architecture.}} The architecture (see Figure~\ref{GANs}) of GANs consists of two neural networks, termed the \emph{generator} and the \emph{discriminator}, which are trained simultaneously through an adversarial process. The architecture can be described as follows:
\begin{itemize}
\item \emph{Generator (G):} The generator network functions as a data synthesizer, taking as input a random noise vector \( z \) from a predefined noise distribution \( p_z(z) \), often Gaussian or Uniform. Its objective is to transform this noise into data \( x \) that are indistinguishable from the genuine data points. The generator's success is measured by its ability to deceive the discriminator.
\item \emph{Discriminator (D):} The discriminator network acts as a binary classifier to distinguish between authentic data drawn from the true data distribution \( p_{\text{data}}(x) \) and synthetic data produced by the generator. Its goal is to accurately identify the source of each data point it evaluates.
\end{itemize}


\textbf{\underline{Mathematical Foundations.}} GANs employ a game-theoretic approach in which two neural networks, the generator (G) and the discriminator (D), engage in a continuous adversarial game. The objective functions for each network define their roles and guide their training. In the following, we detail the mathematical principles that form the basis for GANs.

\emph{Adversarial Objective Function.} The interaction between the generator and the discriminator is mathematically formalized through the following adversarial objective function:

\begin{equation}
    \min_G \max_D V(D, G) = \mathbb{E}_{x\sim p_{\text{data}}(x)}[\log D(x)] + \mathbb{E}_{z\sim p_z(z)}[\log(1 - D(G(z)))] \,.
\end{equation}

Here, \( V(D, G) \) represents the value function for the GANs, which encapsulates the adversarial game:

\begin{itemize}
    \item \textbf{Discriminator's Objective:} The discriminator aims to maximize \( V(D, G) \), by trying to assign the correct label to both real data and generated data. The first term, \( \mathbb{E}_{x\sim p_{\text{data}}(x)}[\log D(x)] \), expects the discriminator to recognize real data (from the dataset) as real, thus maximizing the probability \( D(x) \) for data \( x \) coming from the data distribution \( p_{\text{data}}(x) \).

    \item \textbf{Generator's Objective:} The generator, on the other hand, attempts to minimize \( V(D, G) \) by trying to fool the discriminator. The second term, \( \mathbb{E}_{z\sim p_z(z)}[\log(1 - D(G(z)))] \), measures the generator's success in deceiving the discriminator into believing that the generated data \( G(z) \) are real. Essentially, it tries to minimize the probability that \( D \) recognizes \( G(z) \) as fake.
\end{itemize}

\emph{Equilibrium and Convergence.}
The adversarial training aims to reach a Nash Equilibrium where neither the generator nor the discriminator can improve unilaterally. At this point, the generator produces data that are indistinguishable from real data, and the discriminator is at a loss, unable to differentiate between real and generated data, essentially guessing at a chance level of 50\%.

\emph{Loss Functions and Variants.}
While the original formulation of GANs uses the binary cross-entropy loss, various alternatives and improvements have been proposed such as Wasserstein GAN (WGAN) using Earth Mover's distance for stable learning and meaningful loss metrics~\cite{arjovsky2017wasserstein}, while Least Squares GAN (LSGAN) adopts a least-squares loss function for high-quality generation with reduced mode collapse to stabilize training and resolve issues such as mode collapse \cite{mao2017least}.

\textbf{\underline{Algorithmic Procedure.}} The training of GANs involves an iterative and adversarial process between two distinct networks, the Generator (G) and the Discriminator (D). The following is the refined step-by-step procedure for training GANs:

\begin{enumerate}
    \item {Generate Synthetic Data:} The generator starts by taking a sampled vector of random noise \( z \) from a predefined noise distribution (typically Gaussian). Then it uses this noise as input to generate synthetic data samples \( G(z) \).

    \item {Discriminate Data:} The discriminator examines samples from the real dataset \( x \) alongside the fake data \( G(z) \) produced by the generator. Its goal is to accurately classify real data as real and synthetic data as fake.

    \item {Update Parameters:} 
    \begin{enumerate}
        \item The Discriminator updates its parameters to maximize its ability to correctly classify real and synthetic data. This is often achieved by minimizing the cross-entropy loss for the classification task.
        \item The Generator updates its parameters to better fool the discriminator. It adjusts its weights to minimize the part of the loss function that measures how well the discriminator is able to identify the fake data as fake.
    \end{enumerate}
    Both networks employ backpropagation to update their parameters based on the gradients of the loss function with respect to their parameters.

    \item {Repeat Process:} During training, GANs repeat the above steps for thousands or tens of thousands of iterations, and the generator learns to produce realistic data while the discriminator improves distinguishing real from fake data.
\end{enumerate}

\textbf{\underline{Training and Optimization.}} Training GANs involves careful balancing to prevent either network from overpowering the other, often using techniques such as gradient clipping or modified loss functions to ensure stability. Optimization challenges, such as mode collapse, are addressed through innovations in training protocols and architectural tweaks~\cite{salimans2016improved, arjovsky2017wasserstein,chao2021constrained}.

\textbf{\underline{Extensions and Variants.}} Recent advances have led to various extensions and modifications of the basic GAN architecture, including Conditional GANs (cGANs) for controlled data generation~\cite{mirza2014conditional}, Progressive GANs for improved image synthesis~\cite{karras2017progressive}, and StyleGANs for disentangled representation learning~\cite{karras2019style}. Furthermore, research has focused on improving the stability and convergence of GAN training, such as through the use of techniques such as spectral normalization~\cite{miyato2018spectral} and Consensus Optimization~\cite{mescheder2017numerics}.

\textbf{\underline{Applications.}} GANs have various applications, including: information retrieval~\cite{liang2019unsupervised}, image generation for digital art and entertainment \citep{karras2019style}; data augmentation for medical imaging and other fields \citep{frid2018synthetic}; style transfer for design and art creation \citep{zhu2017unpaired}; super resolution for video enhancement and satellite imaging \citep{ledig2017photo}; image-to-image translation for training simulations and more \citep{isola2017image}; and generating art, pushing the boundaries of creative AI \citep{elgammal2017can}.

\textbf{\underline{Challenges \& Future Directions.}}
Training GANs remains challenging due to several inherent issues, necessitating targeted research to enhance their performance and applicability. One key challenge is \emph{mode collapse}, where the generator produces limited output diversity, reducing the utility of GANs. Strategies such as regularization techniques and diversity-promoting methods like mini-batch discrimination~\cite{salimans2016improved} have shown promise, but future work could further refine these approaches to enhance the robustness of sample diversity.

\emph{Training instability}, arising from the adversarial nature of GANs, often disrupts the convergence and reliability of the model. Gradient penalty techniques~\cite{gulrajani2017improved} and stable architecture designs have mitigated some issues, but future research could explore self-attention mechanisms and adaptive learning rate strategies to achieve more stable and efficient training dynamics.

Another critical issue is the imbalance between the generator and the discriminator. An \emph{overpowering discriminator} can inhibit the learning of the generator, while a \emph{weak discriminator} provides insufficient feedback to make meaningful progress. Current solutions, such as label smoothing and balanced learning rate adjustments~\cite{salimans2016improved}, could be further developed to ensure dynamic equilibrium between the two networks, improving training outcomes.

Beyond these technical challenges, GANs also face significant \emph{computational demands}, which limits their scalability and deployment. More efficient network designs~\cite{wang2020gan} and optimization frameworks could address these limitations, allowing a wider application in resource-constrained environments. Addressing these issues provides a foundation for expanding GAN research into emerging areas.

Future work on GANs aims to extend their capabilities into novel domains. Applications in \emph{3D data generation}, for example, could revolutionize fields like virtual reality, gaming, and medical imaging~\cite{wu2016learning}. Research on ethical implications, including frameworks for the ethical use of GANs and techniques to detect synthetic data~\cite{brundage2021toward}, is critical to ensuring responsible development. Improving \emph{interpretability} of GAN models will promote trust and transparency, making them more suitable for sensitive applications~\cite{berthelot2018understanding}. In addition, leveraging GANs to address social challenges, such as healthcare diagnostics, drug discovery, and climate modeling, highlights their potential for transformative impact.

\textbf{\underline{Summary.}} GANs represent a major breakthrough in AI, fundamentally altering the landscape of machine learning with their novel dual-model architecture. These networks, consisting of a generator and a discriminator, dynamically challenge and advance the traditional boundaries of AI \cite{goodfellow2014generative}. GANs have catalyzed innovation in multiple domains, notably in creating lifelike high-resolution images that redefine the limits of artificial visuals \cite{karras2019style}. They also significantly improve data augmentation, providing critical support in data-sparse fields, and improving the robustness of machine learning models \cite{antoniou2017data}. However, GANs face intrinsic challenges, such as training instability and mode collapse, which highlight the ongoing need for research and development of new methodologies \cite{salimans2016improved}.

\subsection{VAEs}
\label{sec:vae}

\textbf{\underline{Overview.}} Variational Autoencoders (VAEs) represent a class of deep learning models that provide a probabilistic way to describe an observation in latent space. Introduced by~\citet{kingma2013auto} and \citet{rezende2014stochastic}, VAEs are powerful generative models that learn the distribution of data in an unsupervised manner. Their applications range from image generation~\cite{gregor2015draw} and enhancement to complex tasks such as anomaly detection and semi-supervised learning~\cite{an2015variational, kingma2014semi}.

\begin{figure}[!t]
    \centering
    \includegraphics[width=10.5cm,height=5.5cm]{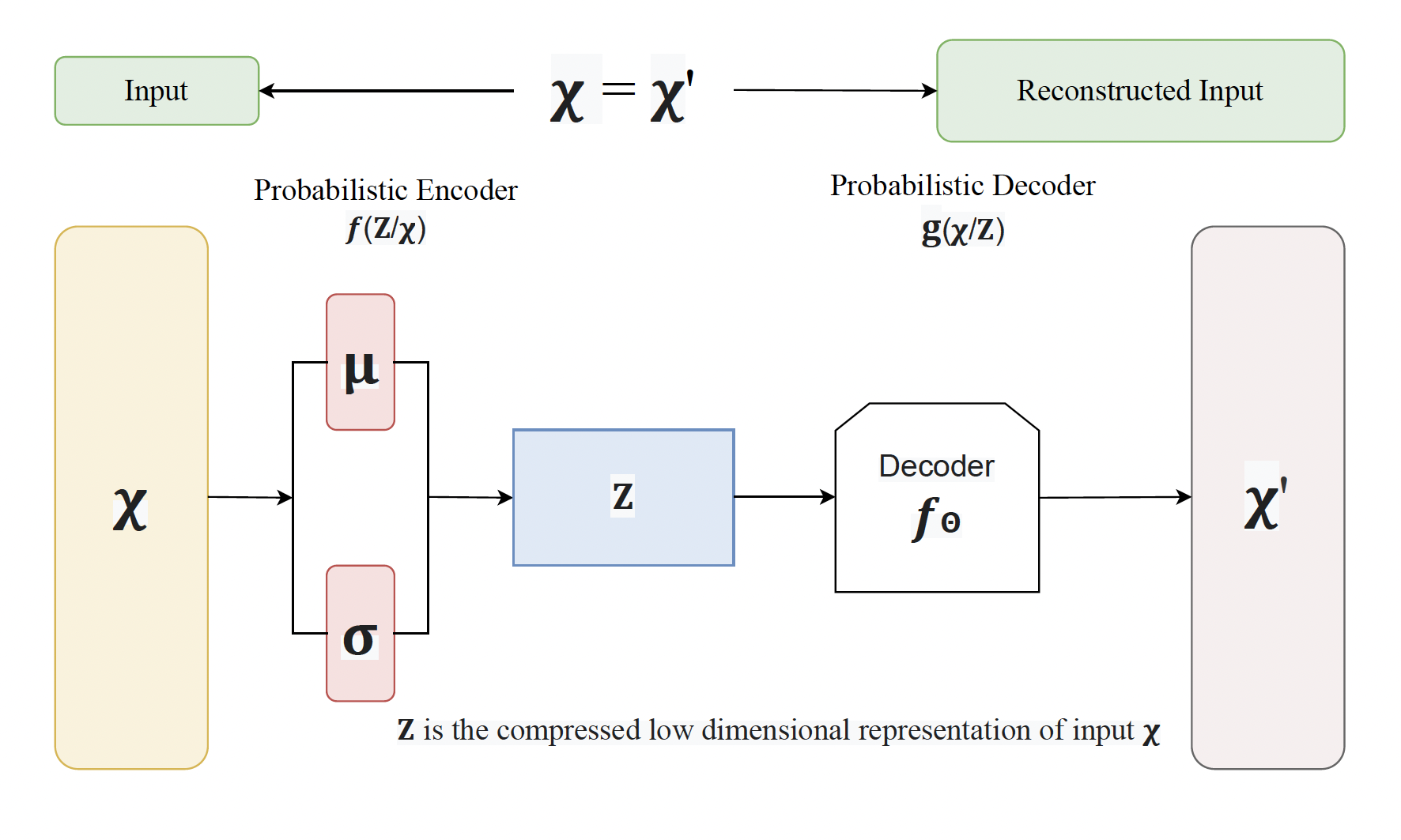}
    \caption{Illustration of VAEs core architecture which consist of an encoder mapping data to latent space and a decoder reconstructing it, jointly trained to optimize reconstruction and latent distribution.} 
    \label{VAE}
\end{figure}

\textbf{\underline{Core Architecture.}} The architecture (see Figure~\ref{VAE}) of VAEs is critical to their functionality, facilitating the learning of efficient data encodings through a probabilistic latent space. The architecture can be described as follows:

\begin{itemize}
    \item \emph{Encoder:} The encoder part of a VAE maps the input data to a distribution over the latent space. Typically, this is achieved through a series of dense or convolutional layers that process the input and output two vectors: one for the mean and one for the variance of a Gaussian distribution~\cite{kingma2013auto}.

    \item \emph{Latent Space:} The latent space is characterized by the mean and variance vectors produced by the encoder. A sample is drawn from this space using the reparameterization trick, which allows the gradient of the loss function to backpropagate through this stochastic step~\cite{kingma2013auto}.

    \item \emph{Decoder:} The decoder uses the sampled latent vector to reconstruct the input data. Similarly to the encoder, this can involve a series of layers that upscale the latent vector back to the dimensionality of the original input~\cite{rezende2014stochastic}.
\end{itemize}

The interaction between these components is mathematically captured by the objective function, which combines reconstruction loss (encouraging the decoder to accurately reconstruct the original data) and the Kullback-Leibler (KL) divergence between the learned latent distribution and the prior distribution, promoting effective and meaningful latent representations.

\textbf{\underline{Mathematical Foundations.}} VAEs are grounded in the framework of probabilistic graphical models with an emphasis on an efficient approximation of the posterior. The mathematical foundations of VAEs are expressed through the following relation:

\begin{equation}
    \log p(x) \geq \mathcal{L}(q) = \mathbb{E}_{q(z|x)}[\log p(x|z)] - D_{KL}(q(z|x) \| p(z))\,,
\end{equation}

\noindent where \(x\) represents the data, \(z\) is the latent variable, \(q(z|x)\) is the approximate posterior, and \(p(z)\) is the prior over the latent variables. The function \(\mathcal{L}(q)\) is known as the evidence lower bound (ELBO), which the VAE optimizes. The first term of the ELBO encourages the decoder to reconstruct the data accurately, while the second term, the KL divergence, acts as a regularizer, enforcing latent variables to approximate the prior distribution~\cite{kingma2013auto, rezende2014stochastic}.

These principles allow VAEs not only to model the data, but also to generate new data instances by sampling from the latent space, making them powerful tools for unsupervised learning in a variety of applications.

\textbf{\underline{Algorithmic Procedure.}} The training of a VAE involves the following steps:

\begin{enumerate}
    \item \textbf{Encoding:} The input data is passed through the encoder, which computes the parameters to the approximate posterior distribution over the latent variables.

    \item \textbf{Sampling:} A latent sample is drawn from the approximate posterior using the reparameterization trick, facilitating gradient backpropagation through stochastic nodes.

    \item \textbf{Decoding:} The decoder uses this latent sample to reconstruct the input data.

    \item \textbf{Loss Computation:} The loss function, which combines reconstruction loss and KL divergence, is computed and minimized during training.

    \item \textbf{Backpropagation:} Gradients of the loss are backpropagated through the network, and the parameters are updated using optimization algorithms such as Stochastic Gradient Descent (SGD) or Adam optimizer.

    \item \textbf{Iteration:} These steps are repeated for multiple epochs over the dataset until the model converges to an optimum.
\end{enumerate}

\textbf{\underline{Training and Optimization.}}
Effective training of VAEs involves several critical techniques to ensure the stability and performance of the model. Key aspects of VAEs training include:

\begin{itemize}
    \item \textbf{Reparameterization Trick:} Essential to enable gradient backpropagation through stochastic variables by expressing the latent variable in terms of a deterministic function of a noise variable~\cite{kingma2013auto}.
    \item \textbf{Variational Inference:} VAEs employ variational inference to approximate the posterior distribution over latent variables, which is crucial to learn efficient latent representations~\cite{blei2017variational}.
    \item \textbf{Learning Rate Scheduling:} Employing techniques such as adaptive learning rate adjustments (e.g., Adam optimizer) enhances convergence and helps avoid local minima~\cite{kingma2014adam}.
    \item \textbf{KL Divergence Annealing:} Gradually increasing the weight of the KL divergence term in the loss function can help to avoid local optima early in training, improving the balance between latent space structure and reconstruction fidelity~\cite{bowman2015generating}.
\end{itemize}

Advanced optimization strategies such as SGD with momentum or the use of second-order optimization methods are also used to refine training dynamics and improve model robustness.

\textbf{\underline{Extensions and Variants.}} Numerous extensions and variants of the basic VAE model have been proposed to address specific challenges and improve performance. These include conditional VAEs, which condition latent space and reconstruction on additional labels or attributes, improving flexibility for tasks such as controlled data generation~\cite{sohn2015learning}. Disentangled VAEs, such as $\beta$-VAE, learn more interpretable and disentangled representations in the latent space, useful for understanding factors of variation in data~\cite{higgins2017beta}. Adversarial VAEs integrate adversarial training, forming Adversarial Autoencoders that regularize the latent space, ensuring diverse and realistic generated samples~\cite{makhzani2015adversarial}.

\textbf{\underline{Applications.}} VAEs have demonstrated their versatility and efficacy in a range of applications, including image generation and reconstruction, anomaly detection, data augmentation, feature learning, drug discovery, user profiling~\cite{liang2021profiling,liang2019collaborative}, and recommender systems~\cite{liang2024survey}. They are widely used to generate new images that mimic the style and characteristics of the training set, useful in graphic design and the creation of digital content~\cite{tufchi2024improved, razavi2019generating}. Using reconstruction probability, VAEs effectively identify anomalies in datasets, which is valuable in surveillance, fraud detection, and maintenance~\cite{xu2018unsupervised, alshameri2024evaluation}. By generating realistic data samples, VAEs enhance the size and diversity of training datasets, which is particularly beneficial in scenarios with limited data~\cite{antoniou2017data, zhang2024vegan}. VAEs are adept at learning meaningful low-dimensional representations of complex data, facilitating feature extraction tasks in domains such as bioinformatics and speech processing~\cite{girin2020dynamical}. In pharmaceuticals, VAEs help to design new molecular structures, which accelerate drug development by efficiently exploring the feasible chemical space~\cite{kadurin2017drugan}.

\textbf{\underline{Challenges \& Future Directions.}}  While VAEs offer significant advantages, they face challenges such as \emph{posterior collapse, limited prior expressiveness, and issues with generalization and overfitting.} Strategies such as KL divergence annealing and modified ELBO objectives can mitigate posterior collapse by improving latent space utilization~\cite{bowman2015generating}. More expressive priors, such as mixture models and autoregressive flows, capture complex data distributions~\cite{tomczak2017vae}, while data augmentation, dropout, and batch normalization improve generalization and robustness~\cite{srivastava2014dropout, shorten2019survey}.

Building on these advances, future work can focus on \emph{efficiency optimization} for resource-constrained deployments, such as mobile devices. Enhancing \emph{variational inference techniques}, which can improve posterior approximations and model stability. Expanding VAEs to \emph{cross-domain applications}, such as audio synthesis and multi-modal learning, opens new possibilities for joint representation learning. Finally, leveraging VAEs for unsupervised and semi-supervised learning and integrating them with \emph{reinforcement learning frameworks} could improve decision-making in complex environments.

\textbf{\underline{Summary.}} VAEs have emerged as a crucial technology in the field of deep learning, offering a robust statistical framework for unsupervised learning through latent space modeling. Introduced by \citet{kingma2013auto} and \citet{rezende2014stochastic}, VAEs harness the principles of Bayesian inference to effectively learn complex data distributions. This approach not only facilitates data generation and reconstruction, but also allows sophisticated applications such as anomaly detection and semi-supervised learning \cite{gregor2015draw, an2015variational}. The reparameterization trick, a key component of VAEs, allows efficient backpropagation and has been instrumental in advancing the training methodologies of generative models \cite{kingma2013auto}. Despite their successes, VAEs face challenges such as posterior collapse and limited prior expressiveness, which can hinder their performance and applicability \cite{bowman2015generating, tomczak2017vae}. Ongoing research is focused on enhancing their architectural efficiency and exploring new applications in fields as diverse as pharmaceuticals and finance, ensuring that VAEs continue to be a cornerstone of generative modeling \cite{sohn2015learning, kadurin2017drugan}. As technology matures, more innovations in variational inference and model optimization are expected to unlock even more potent capabilities, solidifying the role of VAEs in the toolkit of advanced machine learning practitioners.

\subsection{GNNs} 

\textbf{\underline{Overview.}} Graph Neural Networks (GNNs) have emerged as a powerful class of neural networks for modeling structured data as graphs. Introduced in the early 2000s, GNNs have evolved to effectively capture the dependencies in graph-structured data, enabling direct application of deep learning techniques to graphs. These networks are particularly adept at tasks where the data are represented as nodes and edges, such as social networks, molecular chemistry, and knowledge graphs. The ability of GNNs to learn from both node features and graph topology has significantly advanced fields such as node classification, link prediction, and graph classification~\cite{zhou2020graph, wu2020comprehensive, yao2023improving, chen2023heterogeneous}.

\begin{figure}[!t]
    \centering
    \includegraphics[width=11cm,height=6cm]{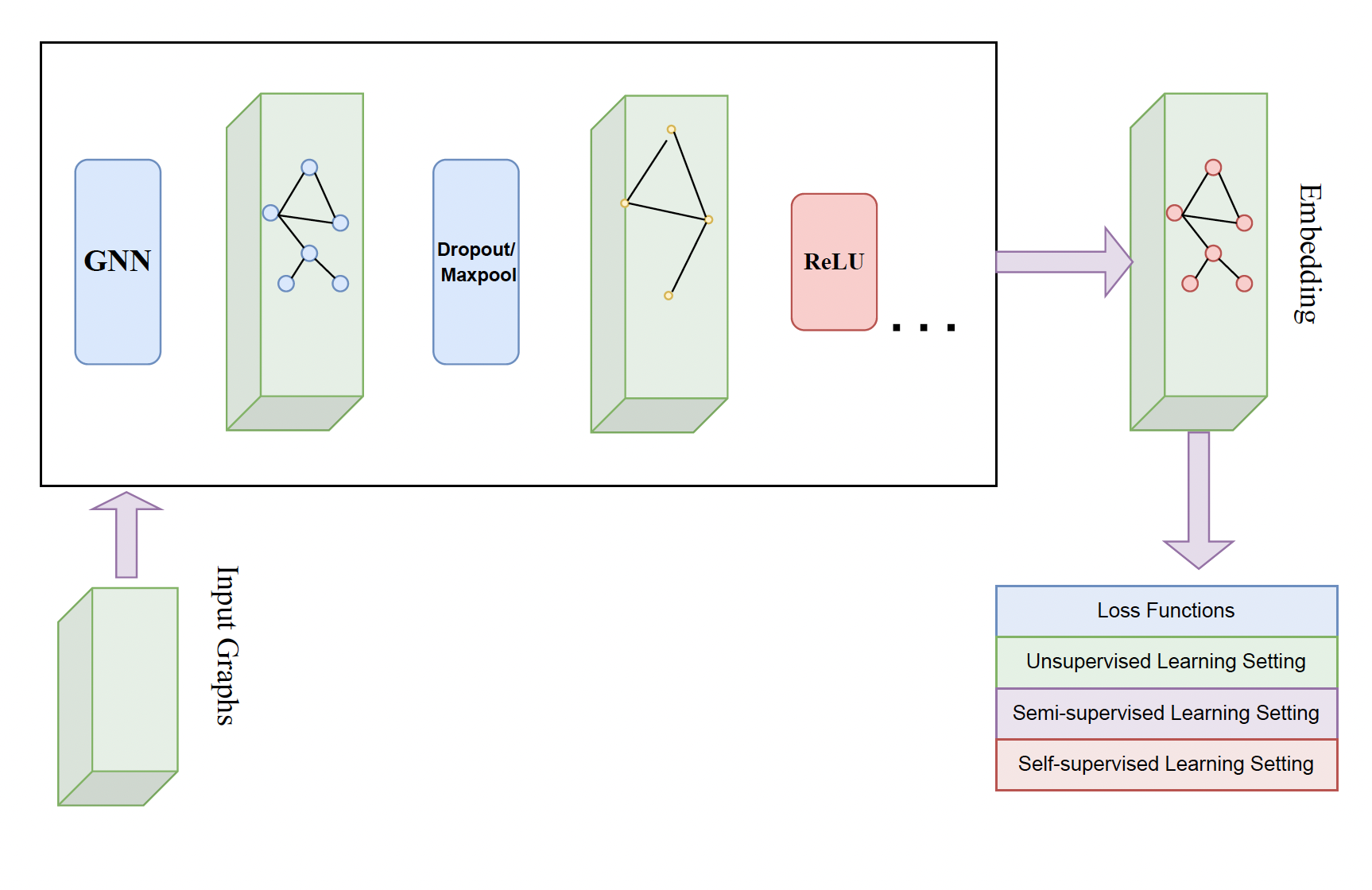}
    \caption{Demonstration of GNNs core architecture where model takes a graph as input (nodes, edges, features) and outputs embeddings for nodes, edges, or the graph. Their architecture involves message passing, where nodes aggregate neighbor features, update embeddings, and perform task-specific predictions using the graph structure.} 
    \label{GNN}
\end{figure}

\textbf{\underline{Core Architecture.}} The foundational architecture (see Figure~\ref{GNN}) of a GNN involves nodes that aggregate information from their neighborhood through message-pass mechanisms. This architecture typically includes layers where each node computes its new state by applying a neural network to aggregate the states of its neighbors, often enhanced by attention mechanisms or gating functions to modulate the information flow. Recent advances have introduced various architectures such as Graph Convolutional Networks (GCNs), Graph Attention Networks (GATs), and more complex models such as Graph Isomorphism Networks (GINs)~\cite{kipf2016semi, velivckovic2017graph, xu2018powerful}.

\textbf{\underline{Mathematical Foundations.}} Mathematically, GNNs are based on the concept of neighborhood aggregation or message passing, where the representation of a node is iteratively updated by aggregating representations of its neighbors. The process is formally defined by:
\begin{equation}
h_v^{(k)} = \sigma\left(W^{(k)} \sum_{u \in \mathcal{N}(v)} \frac{h_u^{(k-1)}}{|\mathcal{N}(v)|} + B^{(k)} h_v^{(k-1)}\right)\,,
\end{equation}
where \( h_v^{(k)} \) is the feature vector of node \( v \) at layer \( k \), \( \mathcal{N}(v) \) denotes the set of neighbors of \( v \), \( W^{(k)} \) and \( B^{(k)} \) are trainable parameters, and \( \sigma \) is a nonlinear activation function. This formulation supports various GNNs models, adapting the aggregation function to incorporate different properties of graphs~\cite{hamilton2017inductive, battaglia2018relational}.

\textbf{\underline{Algorithmic Procedure.}}
\begin{enumerate}
    \item \textbf{Node Feature Initialization:} Each node in the graph is initialized with feature vectors, which could be raw attributes or embeddings.
    \item \textbf{Neighborhood Aggregation:} For each node, aggregate the features from its neighbors using a defined aggregation function, such as the mean, sum or neural attention mechanism.
    \item \textbf{Update Node Representations:} Update each node's feature vector by combining its own features with the aggregated neighborhood features through a neural network.
    \item \textbf{Readout or Pooling:} After several iterations of updates, apply a readout function to derive the entire graph representation or the final features for each node.
    \item \textbf{Train the Model:} Use the derived features for downstream tasks such as classification or regression, and train the model using backpropagation.
\end{enumerate}

\textbf{\underline{Training and Optimization.}} Training GNNs involves several challenges due to the complexity of the graph data. The process is typically optimized for efficiency and effectiveness through techniques such as node sampling, layer-wise training, and using advanced optimizers like Adam or RMSprop. Parallel computing and GPU acceleration are also common to handle large graphs. Recent methods also include graph-specific normalization techniques and dropout strategies to prevent overfitting and improve generalization in various graph-based tasks~\cite{chen2018fastgcn, hamilton2017inductive, rong2020dropedge}.

\textbf{\underline{Extensions and Variants.}} Numerous variants of GNNs have been proposed to address specific challenges such as scalability, dynamic graphs, and heterogeneous graph structures. These include GraphSAGE for inductive learning on large graphs, GATs to introduce attention mechanisms, and more recent developments such as Heterogeneous Graph Neural Network (HetGNN) for dealing with different types of nodes and edges, flow-based GNNs~\cite{liang2021normalizing}. Additionally, there are efforts to combine GNNs with other deep learning architectures such as convolutional and recurrent networks to improve feature extraction and sequence modeling capabilities~\cite{xu2020inductive, velivckovic2017graph, zhang2019heterogeneous}.

\textbf{\underline{Applications.}} GNNs have been successfully applied in various domains reflecting their versatility and high performance. In bioinformatics, GNNs predict protein interfaces and molecular interactions. In social network analysis, they are used for recommendation systems and community detection. GNNs also play a crucial role in traffic networks for route optimization and in financial services for fraud detection. In addition, they are increasingly utilized in natural language processing for document classification and in knowledge graphs for the prediction of entity relations~\cite{gilmer2017neural, fan2019graph, bruna2013spectral, zitnik2018modeling, chen2023heterogeneous}. GNNs have been applied for node alignment in the context of cross-temporal networks~\cite{liang2021cross}, classification~\cite{ling2024bayesian}, drug discovery~\cite{liu2024activelearning}, particle prediction~\cite{lu2024pascl,lu2024factorised}, knowledge graph mining~\cite{cao2024knowledge}. Combining with Gaussian process, GNNs have been applied to relational learning~\cite{fang2021gaussian}.

\textbf{\underline{Challenges \& Future Directions.}}  
While GNNs have demonstrated significant potential, they face several critical challenges that require further research and innovation. One of the primary obstacles is \emph{scalability}, as managing large-scale graphs with millions of nodes and edges imposes substantial computational demands~\cite{chiang2019cluster}. To address this, future studies could focus on developing advanced graph partitioning and sampling methods, such as GraphSAINT~\cite{zeng2020graphsaint}, to optimize computation while preserving the graph structure.

Another key challenge lies in handling \emph{dynamic and heterogeneous graphs}, where evolving topologies and diverse node types complicate representation learning~\cite{kazemi2020representation, zhang2019heterogeneous}. Future research can explore adaptive architectures, such as EvolveGCN~\cite{pareja2020evolvegcn}, and hybrid approaches combining temporal and heterogeneous modeling to enhance the flexibility of GNNs in these contexts.

\emph{Explainability} also remains a significant hurdle, particularly in sensitive applications such as healthcare and finance. Research into interpretability frameworks, such as GNNExplainer~\cite{ying2019gnnexplainer}, can provide clearer insights into how GNNs use node relationships for predictions, thus increasing trust in their output.

Finally, GNNs often face challenges with \emph{generalization}, as they tend to overfit to specific graph structures observed during training~\cite{xu2018how}. To combat this, future work can focus on developing graph-specific regularization techniques, such as DropEdge~\cite{rong2020dropedge}, and customized cross-validation methods to ensure robust performance on unseen data.

\textbf{\underline{Summary.}} Since their introduction, GNNs have become a pivotal algorithm in the analysis of graph-structured data, representing a significant evolution in the field of deep learning. Using the inherent structure of graphs, GNNs efficiently model relationships and interactions in data ranging from social networks to molecular structures, enabling advanced applications such as node classification, link prediction, and graph generation \cite{zhou2020graph, wu2020comprehensive}. The core architecture of GNNs, which combines node feature aggregation with various forms of message passing techniques, has been refined through models such as GCNs, GATs and GINs, enhancing the ability to capture complex dependencies within data \cite{kipf2016semi, velivckovic2017graph, xu2018powerful}. These innovations have led to substantial improvements in predictive performance in various domains, including bioinformatics, recommendation systems, and traffic management \cite{gilmer2017neural, fan2019graph, bruna2013spectral}. Despite these successes, GNNs face ongoing challenges such as scalability, dynamic graph handling, and explainability, which remain active areas of research \cite{chiang2019cluster, kazemi2020representation, ying2019gnnexplainer}. Future developments focus on improving model robustness, reducing computational demands, and extending applications to new, complex problem domains such as climate modeling and computational social science.

\subsection{CLIP}

\textbf{\underline{Overview.}} Introduced by OpenAI in 2021, Contrastive Language-Image Pre-training (CLIP)  employs a novel training methodology designed to learn visual concepts from natural language descriptions. It bridges the gap between vision and language, training on a diverse set of images paired with rich textual descriptions to perform a wide range of visual tasks, such as zero-shot classification. The model's ability to generalize from language nuances to visual representations is a significant advancement, demonstrating broad applicability across various applications~\cite{radford2021learning, goh2021multimodal}.

\begin{figure}[!t]
    \centering
    \includegraphics[width=10cm,height=6cm]{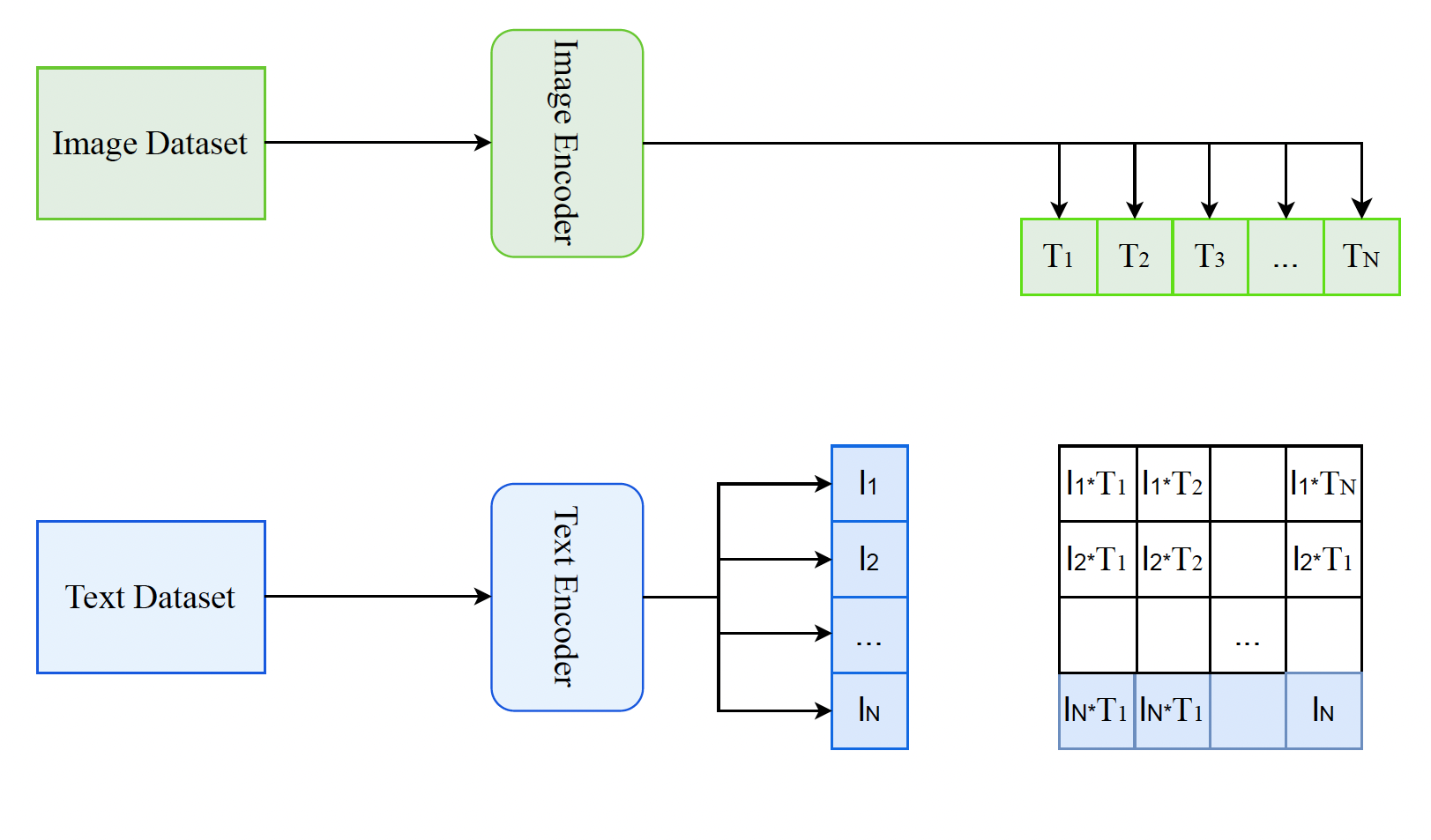}
    \caption{CLIP core architecture processes paired image and text inputs through separate encoders, aligning their embeddings in a shared latent space for tasks like classification and retrieval.} 
    \label{CLIP}
\end{figure}

\textbf{\underline{Core Architecture.}} CLIP architecture (see Figure~\ref{CLIP}) consists of two primary components: a vision transformer and a text transformer. The vision transformer processes images into a series of patch embeddings, while the text transformer encodes textual descriptions into embeddings. Both are trained to optimize a contrastive loss function that aligns the embedding spaces of text and images, enabling effective learning from vast unstructured datasets~\cite{radford2021learning, bao2021beit}.

\textbf{\underline{Mathematical Foundations.}} CLIP's mathematical foundation is centered on maximizing the cosine similarity between correct image-text pairs and minimizing it for mismatched pairs. The contrastive loss function, often referred to as InfoNCE loss, is important to this approach:
\begin{equation}
\mathcal{L} = -\log \frac{\exp(\text{sim}(f(x), g(y)) / \tau)}{\sum_{y' \in Y}\exp(\text{sim}(f(x), g(y')) / \tau)}\,,
\end{equation}
where \( f(x) \) and \( g(y) \) are the image and text embeddings produced by the vision and text transformers, \( \text{sim} \) denotes the cosine similarity, \( \tau \) is a temperature parameter, and \( Y \) is the set of all text candidates~\cite{radford2021learning, wang2021understanding}.

\textbf{\underline{Algorithmic Procedure.}}
\begin{enumerate}
    \item \textbf{Encode Inputs:} CLIP encodes both textual and visual inputs using separate transformer networks. The text encoder processes descriptions into text embeddings, while the image encoder converts images into visual embeddings.
    \item \textbf{Calculate Similarity:} The core of CLIP's learning process involves calculating the cosine similarity between the embeddings of text and the corresponding images within a batch.
    \item \textbf{Contrastive Loss Optimization:} CLIP uses a contrastive loss function (InfoNCE) to align the image and text embeddings by maximizing the similarity of true pairs relative to randomly sampled negative pairs.
    \item \textbf{Backpropagation and Updates:} Based on the calculated loss, the gradients are backpropagated through the image and text networks, and the parameters are updated accordingly to minimize contrast loss.
    \item \textbf{Repeat:} These steps are repeated over multiple epochs using diverse image-text pairs sourced from a wide range of domains and datasets~\cite{radford2021learning, liu2021fusedream}.
\end{enumerate}

\textbf{\underline{Training and Optimization.}} CLIP is trained using a large-scale dataset of image-text pairs. The training process is highly parallelized and optimized for speed and efficiency in modern GPU architectures. Specific optimization techniques include dynamic batching, mixed precision training, and extensive data augmentation to improve the generalization and robustness of the model in various domains and tasks~\cite{radford2021learning, schwartz2020green}.

\textbf{\underline{Extensions and Variants.}} Several extensions and variants of CLIP~\cite{fan2024improving,wang2023exploring} have been explored to enhance its applicability and performance. These include adapting CLIP's approach to other modalities such as audio-visual~\cite{wu2022wav2clip,fan2024revisit,nugroho2023audio} and text-code pairs~\cite{wang2023codet5+}, and integrating it with architectures such as VQ-VAE for complex tasks such as generative modeling. Ongoing research also focuses on modifying its training procedures and loss functions to better handle out-of-distribution data and improve zero-shot capabilities~\cite{radford2021learning, ramesh2021zero, zhou2021denseclip}.

\textbf{\underline{Applications.}}
CLIP has been used effectively in a variety of innovative applications, leveraging its dual understanding of text and images. These include zero-shot learning for image classification without additional task-specific training; multi-modal search engines that can understand and process multi-modal queries~\cite{ramesh2021zero}; content management tools that can understand the context of images and associated texts; automated image captioning generating accurate and relevant captions for images~\cite{radford2021learning}; and educational tools that help create educational content by automatically illustrating complex concepts with relevant images~\cite{patashnik2021styleclip}.

\textbf{\underline{Challenges \& Future Directions.}}
CLIP presents several significant challenges that impact its performance and broader applicability, necessitating targeted research to address these limitations. One critical issue is \emph{bias and fairness,} as the diverse datasets used for training often reflect or amplify existing social biases~\cite{jia2021scaling}. Future efforts can focus on developing methods to detect and mitigate biases during training, ensuring fair and equitable results across applications.

\emph{Data efficiency} poses another challenge, as the large-scale datasets required for training CLIP models can be impractical in data-constrained environments~\cite{radford2021learning}. To address this, research into more efficient learning algorithms and techniques, such as transfer learning and semi-supervised learning, could reduce data dependence while maintaining performance.

\emph{Robustness and safety} remain pressing concerns, particularly in uncontrolled environments or high-stakes applications where failures can have serious consequences~\cite{ramesh2021zero}. Future research could explore techniques to improve robustness against adversarial attacks and unexpected inputs, such as adversarial training or uncertainty modeling.

The \emph{environmental impact} of training large-scale models like CLIP is substantial, given their computational demands~\cite{strubell2019energy}. To mitigate this, innovations in energy-efficient training techniques and hardware optimization, such as the use of green AI practices and advanced processors, could significantly reduce the carbon footprint of model development.

Finally, expanding CLIP’s applications, such as in \emph{robotics and real-time video analysis}, presents an exciting avenue for future work. These extensions require fine-tuning the CLIP’s architecture for specific domains, enabling it to handle dynamic and multimodal data efficiently. By addressing these challenges and following the directions described, future research can enhance CLIP capabilities and broaden its impact in diverse fields.

\textbf{\underline{Summary.}} CLIP represents a significant advancement in the integration of visual and textual data through deep learning. By training on a diverse dataset of images paired with textual descriptions, CLIP effectively bridges the gap between vision and language, enabling a wide range of visual tasks, including robust zero-shot classification \cite{radford2021learning}. The dual-transformer architecture, which combines vision and text transformers, allows effective learning from large unstructured datasets, optimizing through a contrastive loss function that aligns embedding spaces of text and images \cite{bao2021beit}. This has opened up innovative applications from multi-modal search engines to automated image captioning, improving accessibility and user engagement \cite{radford2021learning, ramesh2021zero}. Despite its success, CLIP faces challenges including bias propagation, data efficiency, and ensuring robustness in uncontrolled environments \cite{jia2021scaling, radford2021learning}. The ongoing research aims to address these issues, improve data efficiency, and expand applications, ensuring the adaptability and sustainability of CLIP in future AI deployments \cite{schwartz2020green, zhou2021denseclip}.

\subsection{Diffusion Models}

\textbf{\underline{Overview.}} Diffusion models are a class of generative models that have gained significant attention for their ability to produce high-quality samples. These models simulate a gradual process of adding noise to data and then learning to reverse this process to generate new samples from noise. Since their introduction, diffusion models have been applied to a wide range of tasks, including image and audio generation, demonstrating capabilities that rival those of GANs and VAEs in terms of sample quality and diversity~\cite{sohl2015deep, ho2020denoising}.

\begin{figure}[!t]
    \centering
    \includegraphics[width=0.77\linewidth]{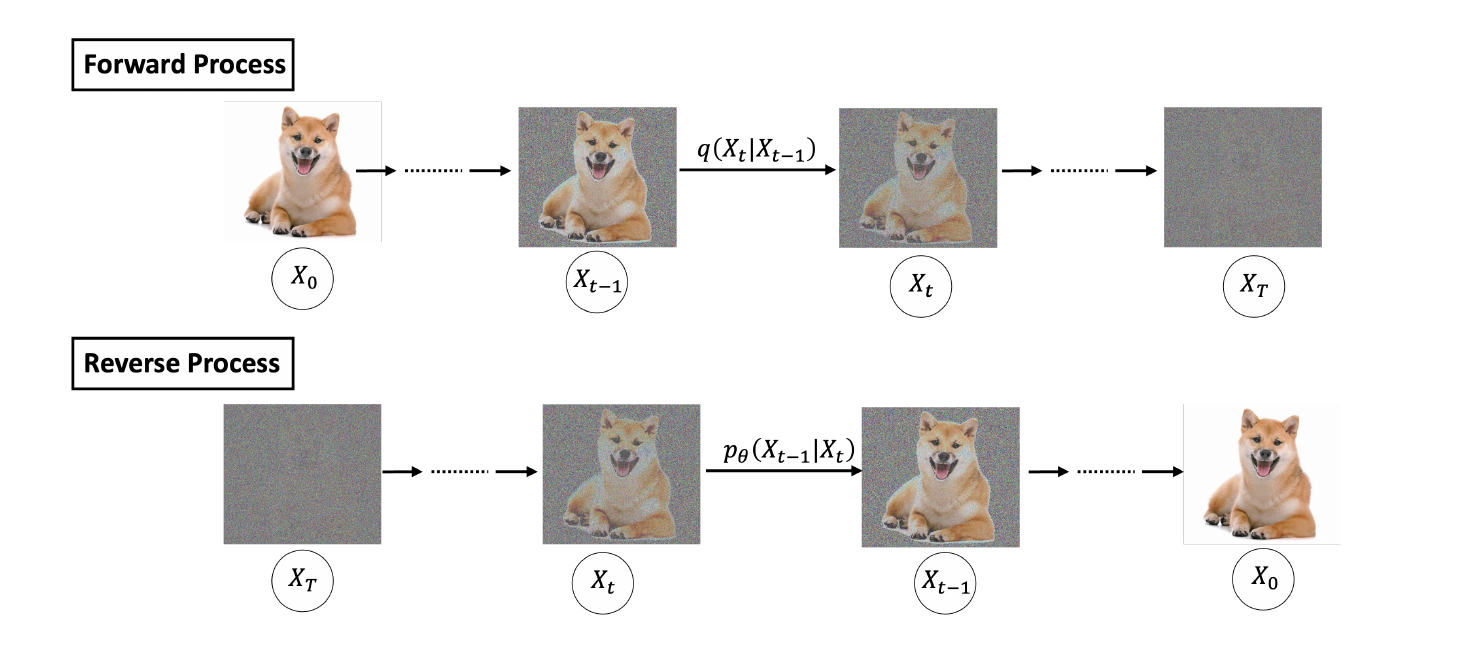}
    \caption{Illustration of diffusion model core architecture where it takes data as input, progressively adds noise to generate noisy samples, and then reverses this process using a neural network to generate data.} 
    \label{dm}
\end{figure}

\textbf{\underline{Core Architecture.}} The core architecture (see Figure~\ref{dm}) of diffusion models is based on a sequence of latent variables that represent the data at various noise levels. This setup involves two main components: the forward process, which incrementally adds Gaussian noise to the data, and the reverse process, a neural network trained to reverse this noise addition. The reverse process is typically parameterized by a deep neural network that predicts noise, conditioned on the noisy data at each step~\cite{ho2020denoising, song2021scorebased}.

\textbf{\underline{Mathematical Foundation.}} Mathematically, diffusion models are grounded in stochastic differential equations (SDE). The forward process can be described by a Markov chain that adds Gaussian noise over a fixed number of steps, while the reverse process involves learning the conditional distribution of the cleaner data given the noisier data. This is often formulated as:
\begin{equation}
p(x_{t-1} | x_t) = \mathcal{N}(x_{t-1}; \mu_\theta(x_t, t), \Sigma_\theta(x_t, t))\,,
\end{equation}
where \( x_t \) is the data at noise level \( t \), and \( \mu_\theta \) and \( \Sigma_\theta \) are learned functions that predict the mean and variance of the reverse Gaussian transition~\cite{song2021scorebased, nichol2021improved}.

\textbf{\underline{Algorithmic Procedure.}}
\begin{enumerate}
    \item \textbf{Forward Process:} Gradually add noise to the data through a predefined number of steps, typically using a variance schedule.
    \item \textbf{Model Training:} Train the reverse model to estimate the conditional distribution of the earlier timesteps given later noisy data.
    \item \textbf{Sampling:} To generate new samples, start with noise and apply the reverse model iteratively to denoise the sample step-by-step until clean data are generated.
    \item \textbf{Refinement:} Optionally refine samples using techniques such as classifier guidance to improve quality or adherence to a target class~\cite{dhariwal2021diffusion}.
\end{enumerate}

\textbf{\underline{Training and Optimization.}} Training diffusion models involves optimizing the parameters of the reverse process neural network, typically using a variant of the variational lower bound. This process can be computationally intensive, often requiring adjustments to learning rates, batch sizes, and noise schedules. Techniques such as dynamic thresholding and learning rate schedules are used to stabilize training and improve convergence rates~\cite{ho2020denoising, kingma2021variational}.

\textbf{\underline{Extensions and Variants.}} Diffusion models have been extended in several innovative ways to enhance their applicability and performance in different domains. Variants like Conditional Diffusion Models allow for controlled generation processes, adapting to specific attributes like class labels or text descriptions. Improvements such as Guided Diffusion and Classifier-Free Guidance integrate additional networks to refine generation toward desired outcomes without explicit conditioning~\cite{croitoru2023diffusion}. Other notable variants include Latent Diffusion Models, which operate in a compressed latent space to enhance efficiency and scalability~\cite{dhariwal2021diffusion, nichol2021improved, rombach2022high}. Techniques similar to those used in autoregressive models for conditional image generation, such as those discussed in~\cite{van2016conditional}, and stabilization methods such as spectral normalization, applicable in various generative model training including diffusion models, as illustrated in~\cite{miyato2018spectral}, further enhance the capabilities and stability of these advanced models.

\textbf{\underline{Applications.}} Diffusion models have shown remarkable success in various domains, including image synthesis, surpassing GANs in fidelity and diversity for tasks like face generation and artistic image creation; natural language processing, contributing to text generation by modeling the distribution of text data; audio synthesis, generating high-quality audio samples comparable to real-world recordings; drug discovery, modeling complex molecular structures and facilitating the design of novel molecules with desired properties; computer vision, employed in tasks such as image-to-image translation, image editing, and image compression; music generation, allowing for the creation of musical pieces that mimic the style of specific composers or genres; efficient data compression, achieving state-of-the-art results in compressing images and videos; and 3D shape generation, enabling the creation of complex 3D shapes with high fidelity. These applications highlight the versatility of diffusion models, demonstrating their capacity to model complex distributions in a variety of data types~\cite{croitoru2023diffusion, ho2020denoising, yang2022diffsound, jing2021torsional}.

\textbf{\underline{Challenges \& Future Directions.}} Despite their impressive success, diffusion models face several significant challenges that require ongoing research to improve their practical deployment and scalability. The \emph{high computational cost} of training and the slow generation process are major barriers. Addressing these issues has led to the exploration of methods such as progressive distillation~\cite{salimans2022progressive} and advanced denoising diffusion models~\cite{ho2020denoising}, which have shown promising results in reducing training times and improving sample quality. Future work could build on these approaches to further optimize efficiency and enable parallelizable algorithms for faster processing.

Another persistent challenge is \emph{sensitivity of diffusion models to hyper-parameters and training configurations}, which requires extensive experimentation and careful tuning. Future research could focus on automating hyper-parameter selection and developing robust training strategies to enhance stability and simplify the training process. Ensuring the diversity and quality of the generated samples while avoiding mode collapse remains an important area of improvement~\cite{song2021scorebased, ho2020denoising}. Hybrid models that combine diffusion with other generative frameworks could offer solutions by leveraging complementary strengths to improve sample quality and training robustness.

Furthermore, the need to better understand the underlying mechanisms of diffusion models, including their \emph{optimization dynamics and convergence properties}, presents both a challenge and an opportunity. Advances in this area could not only improve theoretical understanding, but also guide the design of more efficient models~\cite{dhariwal2021diffusion}. Exploring these dynamics in broader contexts, such as video generation, 3D data synthesis, and multi-modal applications, remains a promising direction for future work.

Recent developments in applying diffusion models to tasks such as image-to-image translation~\cite{yang2022diffusion} and text-to-image synthesis~\cite{rombach2022high} highlight their versatility. Building on these advancements, the development of efficient architectures, such as Efficient Diffusion Models (EDMs)\cite{salimans2022progressive} and Hierarchical Diffusion Models (HDMs)\cite{liu2022Hierarchical}, demonstrates the potential to achieve state-of-the-art results in compressing images and videos. Future work can refine these approaches to expand their application across diverse domains while improving scalability and reducing resource demands.

\textbf{\underline{Summary.}} Introduced in the realm of generative modeling, diffusion models have quickly risen to prominence for their exceptional ability to produce high-quality samples across various data types. By mimicking the natural process of diffusion through an addition of forward noise followed by a reverse denoising path, these models have set new standards in image and audio synthesis, rivaling and sometimes surpassing the capabilities of GANs and VAEs \cite{sohl2015deep, ho2020denoising}. The core architecture of diffusion models, characterized by a sequence of transformations that gradually revert noisy data to their clean state, leverages deep neural networks to predict and reverse noise, grounding its operations in stochastic differential equations for precise control over the generative process \cite{ho2020denoising, song2021scorebased}. The applications of diffusion models extend beyond the mere generation of images, impacting fields such as natural language processing and drug discovery by modeling complex, high-dimensional data distributions \cite{ho2020denoising, yang2022diffsound, jing2021torsional}. Despite these advancements, diffusion models face challenges, including high computational demands and slow sample generation times, necessitating ongoing research to enhance their efficiency and applicability \cite{watson2021learning, kingma2021variational}. Future research directions are exploring hybrid models that combine the strengths of diffusion with other generative approaches, in order to improve both the output quality and the stability of the model \cite{hojonathan2021cascaded, song2021scorebased}.

\section{Discussion}
\label{sec:discussion}

This section explores core components of deep learning, including building blocks, normalization techniques, optimization algorithms, and rate scheduling, and connects them to specific models: ResNets, Transformer, GANs, VAEs, GNNs, CLIP, Diffusion Models.

\subsection{Core Components and Techniques}
\textbf{Building Blocks.} CNNs extract spatial hierarchies, enhanced by residual connections \cite{he2016deep}, as seen in ResNets, while encoder-decoder frameworks drive models like Transformers \cite{vaswani2017attention}, extending to vision and graph domains \cite{ying2021transformers}.

\textbf{Normalization.} Layer Normalization stabilizes training in Transformers \cite{xiong2020layer} and VAEs \cite{dziugaite2020revisiting}, while Batch and Instance Normalization address stability in ResNets \cite{ioffe2015batch} and style transfer tasks \cite{huang2017arbitrary}.

\textbf{Optimization and Scheduling.} ADAM \cite{kingma2014adam} and RMSProp \cite{tieleman2012lecture} are critical for models such as GANs \cite{radford2015unsupervised} and Transformers, supported by learning rate schedulers such as cosine annealing \cite{loshchilov2016sgdr}.

\subsection{Integration with Models}
\begin{itemize}
    \item \textbf{ResNets}: Residual connections and Batch Normalization enable deep networks for image recognition \cite{he2016deep}.
    \item \textbf{Transformers}: The encoder-decoder architecture with layer normalization, ADAM, and cosine scheduling effectively captures long-range dependencies \cite{vaswani2017attention}.
    \item \textbf{GANs}: Employ RMSProp, ADAM, and rate scheduling to stabilize adversarial training, excelling in generative tasks \cite{brock2019large}.
    \item \textbf{VAEs}: Leverage Layer Normalization and ADAM to optimize latent space representations for applications such as image generation \cite{kingma2013auto}.
    \item \textbf{GNNs}: Message passing with normalization and ADAM supports structural learning in graph data \cite{kipf2016semi}.
    \item \textbf{CLIP}: Combine ADAM and normalization for large-scale multi-modal learning, bridges vision and language \cite{radford2021learning}.
    \item \textbf{Diffusion Models}: Use normalization, advanced optimization, and scheduling to model data generation processes effectively \cite{ho2020denoising}.
\end{itemize}

\section{Conclusion \& Future Work}
\label{sec:conclusion}
This survey provides an overview of seven deep learning algorithms that have transformed AI over the past decade. We break down every algorithm, exposing its workings and special qualities. We designed our survey as a practical guide to help experienced researchers transitioning into deep learning from other domains, as well as beginners seeking to understand trending algorithms. In addition, our study fosters further exploration by emphasizing the importance and immense potential of these algorithms.

In future work, we plan to expand our scope to include more algorithms, investigate their applications in various domains, and provide a deeper analysis of their strengths and limitations. In addition, our goal is to explore the development of new evaluation measures and benchmarks to further advance the field of DL.

\section*{Limitations}

While we have made every effort to provide a comprehensive overview of the seven deep learning algorithms, certain areas present opportunities for further improvement and exploration. First, certain contributions may have been overlooked due to differences in terminology across studies and communities, despite our efforts to include diverse perspectives and relevant citations. To mitigate this, we conducted a thorough literature search and cross-referenced multiple sources to ensure broad coverage. Furthermore, the future research directions proposed for each algorithm are necessarily limited in scope. Some directions may already have been explored, while others could benefit from deeper refinement and specificity. Addressing these limitations will involve methodological improvements, including more dynamic approaches to capture advancements and ensuring the comprehensive inclusion of related contributions.

\section*{Ethical Statement}
We conducted the survey with participants who have informed consent for their data and responses to be used for research purposes and potential publication. To protect their identities, all participants remained anonymous throughout the study, and we deliberately avoided asking questions that could compromise their anonymity.

\section*{Bias}

Our methodology is based on survey responses, and the selection of algorithms can reflect biases arising from the experience and background knowledge of the respondents. In particular, most of the respondents were affiliated with a limited number of universities, potentially introducing institutional biases. Despite our substantial efforts to distribute the survey widely among professionals from diverse institutions and backgrounds, there is room to further enhance representation. Expanding the participant pool to include a broader and more globally representative group is a critical direction for future work. The targeting of underrepresented regions, industries and interdisciplinary researchers can further enrich the diversity of perspectives and reduce potential biases in future studies.

\section*{Acknowledgment}

The authors thank Bhaskar Mukhoty and Zaigham Zaheer for their invaluable suggestions and proof-reading of our manuscript. 

\section*{Conflict of Interest}
The authors declare that they have no conflicts of interest related to this work.

\clearpage

\bibliography{main}
\bibliographystyle{tmlr}


\end{document}

%% file: math_commands.tex
\usepackage{amsmath,amsfonts,bm}

\def\eqref#1{equation~\ref{#1}}

\def\1{\bm{1}}

\DeclareMathAlphabet{\mathsfit}{\encodingdefault}{\sfdefault}{m}{sl}
\SetMathAlphabet{\mathsfit}{bold}{\encodingdefault}{\sfdefault}{bx}{n}

